\documentclass[journal]{vgtc}
\usepackage[utf8]{inputenc}
\usepackage{multirow}

\ifpdf%                                % if we use pdflatex
  \pdfoutput=1\relax                   % create PDFs from pdfLaTeX
  \usepackage{graphicx}                % allow us to embed graphics files
  \DeclareGraphicsExtensions{.pdf,.png,.jpg,.jpeg} % for pdflatex we expect .pdf, .png, or .jpg files
\else%                                 % else we use pure latex
  \usepackage{graphicx}                % allow us to embed graphics files
  \DeclareGraphicsExtensions{.eps}     % for pure latex we expect eps files
\fi%

\graphicspath{{figures/}} % where to search for the images
\usepackage{microtype}                 % use micro-typography (slightly more compact, better to read)
\PassOptionsToPackage{warn}{textcomp}  % to address font issues with \textrightarrow
\usepackage{textcomp}                  % use better special symbols
\usepackage{mathptmx}                  % use matching math font
\usepackage{times}                     % we use Times as the main font
\usepackage{cite}  

\usepackage{comment}
\usepackage[dvipsnames]{xcolor}

\usepackage[normalem]{ulem}  % for Strikethrough text

\usepackage{tabu}                      % only used for the table example
\usepackage{booktabs}                  % only used for the table example
\usepackage{xcolor}
\usepackage{blindtext}
\vgtccategory{Research}
\vgtcpapertype{algorithm/technique}

\newcommand{\change}[1]{#1}

\title{ErgoExplorer: Interactive Ergonomic Risk Assessment\\
from Video Collections
}
\author{Manlio Massiris~Fernández, Sanjin Rado\v{s}, Kre\v{s}imir Matkovi\'{c}, M. Eduard Gr\"{o}ller, and Claudio Delrieux }
\authorfooter{
\item
 Manlio Massiris~Fernández is with Departamento de Ing. Electrica y Computadoras, Universidad Nacional del Sur, Escuela de Ingenierías Industriales, Universidad de Extremadura, and CONICET. E-mail: mmassiris@uninorte.edu.co.
 \item
 Sanjin Rado\v{s} and Kre\v{s}imir Matkovi\'{c} are with VRVis Research Center in Vienna. E-mail: Rados, Matkovic@VRVis.at.
 
 \item
 M. Eduard Gr\"{o}ller is with TU Wien. E-mail: meister@cg.tuwien.ac.at.
\item
 Claudio Delrieux is with Departamento de Ing. Electrica y Computadoras, Universidad Nacional del Sur, and CONICET. E-mail: cad@uns.edu.ar.
 }

\shortauthortitle{Massiris Fernandez \MakeLowercase{\em et al.}: Visualization of Ergonomic Assessment}

\abstract{
Ergonomic risk assessment is now, due to an increased awareness, carried out more often than in the past. 
The conventional risk assessment evaluation, based on expert-assisted observation of the workplaces and manually filling in score tables, is still predominant. 
Data analysis is usually done with a focus on critical moments, although without the support of contextual information and changes over time. 
In this paper we introduce {\tt ErgoExplorer}, a system for the interactive visual analysis of risk assessment data. 
In contrast to the current practice, we focus on data that span across multiple actions and multiple workers while keeping all contextual information. 
Data is automatically extracted from video streams. 
Based on carefully investigated analysis tasks, we introduce new views and their corresponding interactions. 
These views also incorporate domain-specific score tables to guarantee an easy adoption by domain experts. 
All views are integrated into {\tt ErgoExplorer}, which relies on coordinated multiple views to facilitate analysis through interaction.
{\tt ErgoExplorer} makes it possible for the first time to examine complex relationships between risk assessments of individual body parts over long sessions that span multiple operations. 
The newly introduced approach supports analysis and exploration at several levels of detail, ranging from a general overview, down to inspecting individual frames in the video stream, if necessary. 
We illustrate the usefulness of the newly proposed approach applying it to several datasets.
}


\teaser{
  \centering
\includegraphics[width=\linewidth]{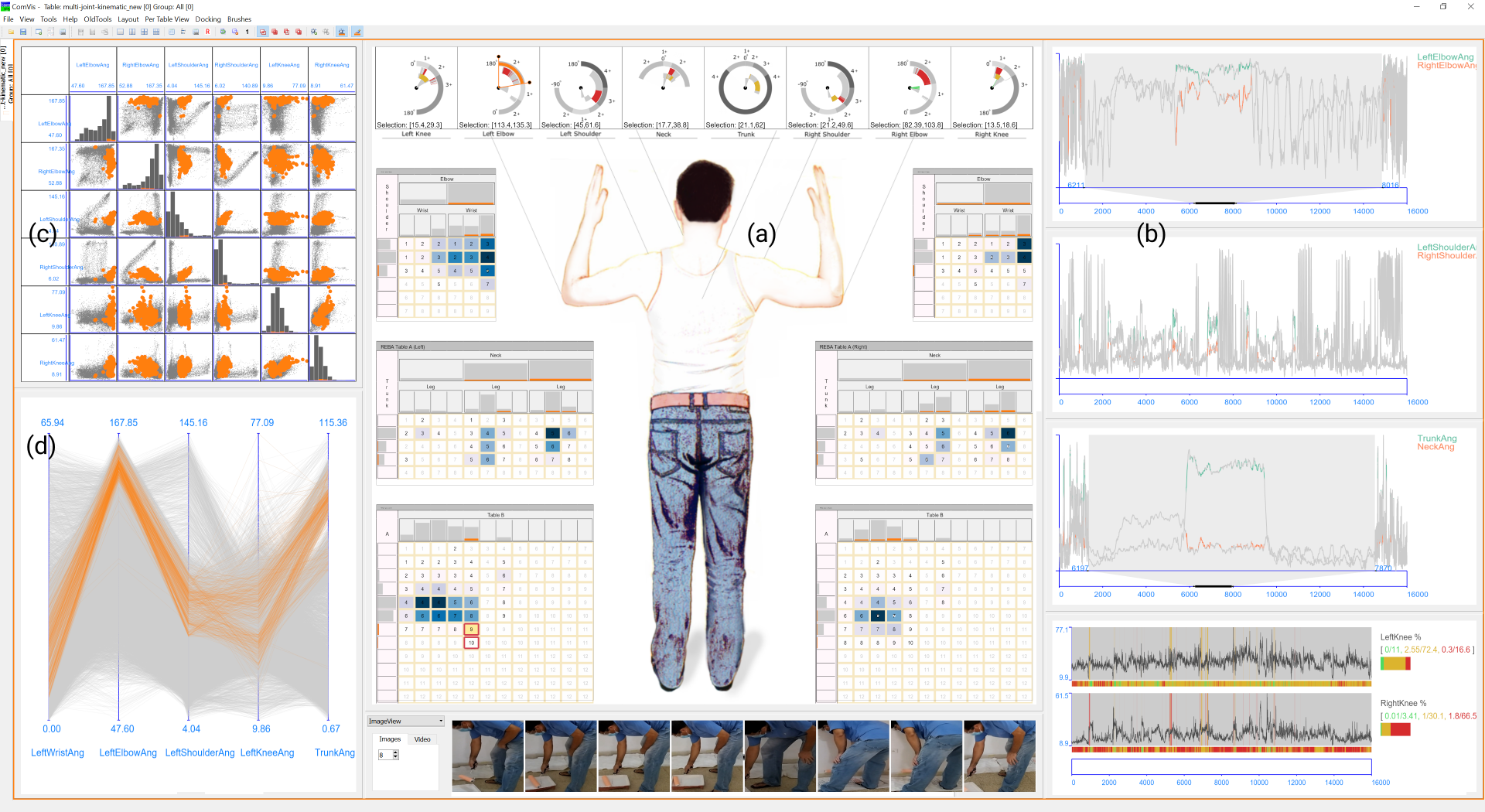}
  \caption{A dashboard to support the analysis of complex relationships between risk assessments of individual body parts over long sessions that span single or multiple operations. Ergonomists can understand how to mitigate ergonomic risk by using coordinated multiple views (CMV) \textbf{(a)} {\tt ErgoView}, \textbf{(b)} {\tt ErgoTimeline}, \textbf{(c)} scatter plot matrix, and \textbf{(d)} parallel coordinates.
  }
	\label{fig:teaser}
}

\begin{document}

\maketitle

\section{Introduction}

Recent advancements in smart factories, via Industry 4.0 (I4.0), raised generalized requirements concerning a more thorough analysis of workers' activities in several settings. These include manufacturing, process industries, and construction, to mention just a few~\cite{Manghisi2020}.
Also, workplace accidents or other health-related incidents that might cause injuries to workers, raise legal disputes in which carefully collected evidence may be required to set out the actual responsibilities and eventual compensations~\cite{FAN201385}.
However, traditional workplace activity monitoring and ergonomic assessments rely on self-reporting or specialists' direct observations.
In risk evaluation, for instance, observations regarding the human body focus on measuring angles of the trunk and limb joints. 
This makes the procedure costly and severely prone to intra- and inter-observer variances~\cite{Schwartz2019}.
Several alternatives to automate procedures for tracking and collecting ergonomic data have been proposed, including accelerometers, RFID devices, motion sensors, LiDAR scanners, GPS, physiological monitoring, and many others.
We refere to Subedi and Pradhananga~\cite{Subedi2021} for a thorough literature review on the topic.

The new data-driven I4.0 framework requires a more systematic and unbiased collection and analysis of workers' activities. This helps optimize processes in manufacturing, construction, and various other industrial settings, in which human-assisted observation is inadequate.
Computer Vision (CV) is becoming the main alternative to human-assisted monitoring. Recently several CV-based approaches have been proposed aimed to provide adequate unsupervised solutions to ergonomic assessment.
In particular, deep-learning based body pose estimation like STAF~\cite{Raaj2019} or VIBE~\cite{Kocabas2019} are enabling novel and significant breakthroughs in several contexts, including workspace ergonomic assessment~\cite{MassirisFernandez2020a}.
This trend enabled a complete digitization of the different workplace aspects, including personnel activities, providing a wealth of potentially valuable data.
In ergonomic assessment, for instance, routine monitoring may require evaluating about 25 joint angles and their combined relative values at least once per second.
Thus, a single worker's hourly activity generates a large amount of data that require adequate tools to perform exploratory analyses.
This helps ergonomists pinpoint situations or contexts that require intervention or find relevant situations that may inform litigation.
However, to the best of our knowledge, there are no attempts to provide means to extract and handle significant ergonomic information in workplaces in a sensible way.

In this work we introduce {\tt ErgoExplorer}, a visualization tool able to explore and analyze time dependent ergonomic scores from observations encompassing very long periods.
By means of CV-based data extraction from regular cameras, the system can analyze and synthesize large amounts of ergonomic data. The goal is to facilitate the ergonomists' tasks to detect unwanted or unexpected workplace conditions.
This in turn enables insights from complex interplays of situations, to evaluate possible workplace scenarios on a sound basis, and to address matters in which a careful yet massive ergonomic data analysis must be carried out.
{\tt ErgoExplorer} is based on coordinated multiple views, incorporating the traditional score tables used by ergonomists in an interactive manner.
They are linked with other depictions, including two novel views, {\tt ErgoView} and {\tt ErgoTimeline}.
In addition we customize other views to study temporal aspects of the scores.

% In Section~\ref{sec:Background} we briefly introduce the required background to make this paper self contained.
% In Section~\ref{sec:Previous_Work} we present the relevant prior work and discuss the differences to previous assisted or unsupervised assessment of ergonomic risk proposals and their limitations.
% \textcolor{red}{
% Section~\ref{sec:Requirement_Analysis} elicits... introduces the proposed approach.\\
% A  visual analytics system design for our ErgoExplorer is then introduced in Section 5.
% The scientific novelites of this paper are...\\
% a novel task abstraction for ergonomic task analysis} 

\section{Ergonomic Assessments}\label{sec:Background}

Workplaces enforcing a safety culture are known to be more productive, with high employee morale and low burnout.
A careful ergonomics integration is among the most relevant factors in this enforcement.
Work-related musculoskeletal disorders (WMSDs) are typical workplace health issues that may result in inflammation or degeneration of functional body structures~\cite{Ha2009}.
WMSDs are the leading cause of productivity losses due to deaths and work-related permanent disabilities, litigation, sick leaves, and the related indirect costs~\cite{Bevan2015a}.
Ergonomic assessment programs aim to detect and assess actual or potentially harmful workplace situations, suggesting interventions to prevent the occurrence of WMSDs~\cite{Luttmann2004}.

Traditional workplace activity monitoring and ergonomic assessments rely on self-reporting or on specialists' direct observations regarding the angles of main body joints.
Rapid Upper Limb Assessment (RULA)~\cite{McAtamney1993} and Rapid Entire Body Assessment (REBA)~\cite{McAtamney2000} appear to be the most widespread approaches in this respect~\cite{Lowe2019}.
Human-assisted evaluation has to be performed by specialized ergonomists.
The complexity of such a manual task typically requires the evaluation to focus only on critical moments in which the workers are executing potentially or actually risky movements.
With pencil-and-paper-based methodologies, ergonomists observe workers performing tasks and then fill a table.
They are slow and cumbersome when it comes to monitoring a worker over an extended period of time.
This requires large intervention times of human experts, whose sole presence may also alter the actual worker's performance. 
As a consequence, the resulting assessments are vulnerable to observer fatigue, have scalability difficulties, and depend heavily on the experts' subjective criteria.
This leads to non-uniform evaluations, intra- and inter-observer variance, and other detrimental issues~\cite{Schwartz2019,Roberts2020}.
However, the scoring tables are essential as experts rely on them to consolidate the measurements taken throughout an activity and they are the primary tool for decision making up to now.
% Also, typical situations that may incur ergonomic risk, for instance movement repetitions or prolonged standing postures, are even more difficult to assess by direct human observation.
% Finally, I4.0 is imposing constant data-driven analyses and improvements to meet the demands of global competitive markets and sustainable practices in an efficient manner.
% In particular, workplace monitoring is one of the main data sources of key factors that may affect efficiency.

%\change{\subsection*{Rapid Entire Body Assessment}}%\tt 
%\label{REBA method}

% In the context of ergonomic data analysis, evaluation tables such as those used by REBA~\cite{McAtamney2000} have so far played a central role.
% and have been widely accepted by ergonomists \cite{Lowe2019}.
\change{\textbf{Rapid Entire Body Assessment} (REBA) generates a postural examination framework sensitive to musculoskeletal risks in various work tasks.
It is specifically applicable to unpredictable working postures found in construction, health care, and other service industries~\cite{McAtamney2000}.
With REBA, body joints ({\em e.g.}, shoulders or elbows) are evaluated based on their angular deviations from a predefined safe and comfortable posture. 
Ergonomists assign individual joint and posture scores to each body region for a visually monitored working task.
Depending on the specific case, they fill  additional worksheets for each significant change of a body posture.
The ergonomic assessment considers bio-mechanical and postural load requirements on the neck, trunk, and upper limbs during the work cycle.
For this, a systematic process has to be performed to evaluate the required body postures, exerted forces, and repetitions for the tasks being assessed.
Moreover, limbs are analyzed separately (e.g., right versus left upper arm) if performing different actions.
So instead of a single-page worksheet containing the three tables, ergonomists need additional tables to assign respective scores.
Deviations from the predefined safe postures receive individual scores that are integer numbers related to the actual angular deviations in all joints.
Joint scores are then combined into limbs and trunk scores using specific tables typically presented in {\em ad-hoc} worksheets.
The limbs and trunk scores finally guide the computation of the overall grand score}.

% using the latter figures.
%As already mentioned, the current trend to automate these procedures takes advantage of breakthroughs in computer vision.
%Traditional feature-engineered CV procedures typically face the perspective distortion problem in assessing joint angles, and thus muti-camera data fusion to track the operators during their work was frequently employed~\cite{Bauters2017}.
%The joint angles are then calculated by means of a 3D-model of the worker using the synchronized multiple cameras footage.
%This requires very complex {\em ad-hoc} background removal techniques and geometric analyses.
%More recently, open-source real-time systems for multi-person key-point detection libraries were developed using deep learning non-parametric methods.

\section{Previous Work}\label{sec:Previous_Work}
Visual analysis of joint movements was proposed in several contexts~\cite{Bernard2017}, including the study of locomotion~\cite{Whittle2006, Lu2020a}, deviations in movement patterns~\cite{Manal2005}, ortopedics-oriented biokinematic data~\cite{Krekel2010,Chan2019}, sports~\cite{Kanazawa2019}, and human-computer interaction~\cite{Yeo2013}.
However, there are only a few proposals in the literature targeting the visual analysis of workers' performance in actual workplace settings.
The work of Han~et~al.~\cite{Sang2011} introduces simulated 3D working environments to perform ergonomic analyses.
The simulated environment is used as an observation tool.
Specialists can assess the different tasks and activities that the workers will perform, and if they will result in unhealthy postures, inappropriate repetitions, excessive force or static loading, or stress in some body parts.
Teizer~et~al.~\cite{Teizer2013} present an educational and training environment for construction workers that is based on the integration of real-time location tracking and immersive 3D data visualization. 
Kanan~et~al.~\cite{Kanan2018} propose an autonomous system to monitor the position of workers and equipment.
Its effectiveness has been tested on real construction sites. 
As a result, they display a video summary to supervise the proximity of the worker and the risk area around the equipment. 

More automatized proposals apply CV-based ergonomic analyses.
Bauters~et~al.~\cite{Bauters2017} discuss a multicamera-based model to automate the analysis of assembly workstations.
The proposed system generates real-time information to support improvements in workers' performances.
The collected data is analyzed in terms of specific key performance indicators (KPIs).
The resulting information is presented in a specifically designed dashboard that visualizes the workers' efficiency, pace, value-adding activities, anomalous work cycle situations, and other relevant parameters.
Video processing is based on traditional engineered feature extraction.
Thus, the approach is less flexible compared to state-of-the-art practices, and highly dependent on the fine-tuning of several processing aspects, camera settings, and data fusion.

\begin{table*}[t]
\caption{Task types.}
\label{table_ttt}
\begin{tabular}{|l|l|l|}
\hline
\textbf{Ergonomic Task Analysis}  
& \textbf{Typical Topic Question}  
& \textbf{Abstract Task}\\ \hline

\multirow{2}{*}{T1: Determine the type of ergonomic analysis} 
& Q1: What is an adequate overview? 
& Categorize (Fig~\ref{fig:teaser}) \\
\cline{2-3}
& Q2: What is the purpose of the analysis? 
& Distinguish \\
\hline

\multirow{2}{*}{T2: Define goals and evaluation criteria} 
& Q3: What are the expected outcomes of this analysis? 
& Identify \\ 
\cline{2-3} 
& Q4: What are the performance assessment criteria?     
& Compare (Fig~\ref{fig:teaser}) \\ \hline

\multirow{4}{*}{T3: Task decomposition table or diagram} 
& Q5: How can this work be split into tasks? 
& Distinguish\\
\cline{2-3} 
& Q6: How many times is each task performed? 
& Compare \\ 
\cline{2-3} 
& Q7: How much time is spent on each task?
& Compare\\
\cline{2-3} 
& Q8: Which task should be analyzed first? 
& Identify\\ \hline

\multirow{2}{*}{T4: Check decomposition validity} 
& Q9: Which data represents the task best?
& Locate\\ 
\cline{2-3} 
& Q10: Are outliers and wrong data filtered out? 
& Distinguish\\ \hline

\multirow{3}{*}{T5: Identify risky movements} 
& Q11: What risky movements are related to a particular task?
& Identify \\ 
\cline{2-3} 
& Q12: How are scores distributed in the risk tables? 
& Categorize\\
\cline{2-3} 
& Q13: Which body joints present high risk? 
& Distinguish\\ 
\cline{2-3}
& Q14: Is the risk balanced on both sides of the body? 
& Compare\\ 
\hline

\multirow{3}{*}{T6: Test hypotheses concerning performance factors} 
& Q15: When is intervention required?   
& Distinguish \\
\cline{2-3} 
& Q16: How to mitigate ergonomic risk?     
& Identify \\
\cline{2-3} 
& Q17: How to validate improvements?   
& Compare\\ \hline
\end{tabular}
\label{Tab:Ergonomic Task Analysis Steps}
\end{table*}

A similar goal was investigated by Li~et~al.~\cite{Li2018}, where the authors apply 3D skeletal modeling to emulate the workers' movements in real construction and manufacturing sites.
According to the authors the method is able to discern ergonomic risks by detecting inadequate body postures and also to evaluate force and load handling that may potentially generate injuries.
The actual information involved in tasks maneuvering have to be delivered from the workplace design and environment, and the task schedule, which, according to the authors, can be obtained from direct observation or video recordings.
The resulting data is then used together with the mentioned 3D model to infer ergonomic information such as the joints’ locations and angles.
This enables a successive risk assessment analysis using traditional methods like RULA~\cite{McAtamney1993} or REBA~\cite{McAtamney2000}.

The latest breakthroughs in deep learning applied to CV enable a more thorough and rigorous monitoring, in particular by means of body-pose estimation-modules like STAF~\cite{Raaj2019} or VIBE~\cite{Kocabas2019}.
In~Massiris~Fernandez~et~al.~\cite{MassirisFernandez2020a}, for instance, the effectiveness and accuracy of a CV-based approach was tested in a variety of scenarios.
Difficult workplace settings with several workers are included, involving occlusions and self-occlusions, varying illumination conditions, moving and egocentric cameras, etc.
The success of this approach triggers new challenges, in particular how to convey and help make sense of the large amount of activity data that is collected second by second.

Another significant aspect of video-based massive data collection is related to adequately filter out irrelevant parts of takes that may hamper the significance of the overall data analysis and visualization.
Workers often perform movements that are unrelated to their actual activities, and in a different time granularity.
Examples include a short arm movement to scratch the forehead, a more prolonged arm and head movement to check the time on a smartphone, an even longer and more complex sequence of movements to grab a water bottle, and sip, etc.
Indiscriminate movement collection and analysis, oblivious to the aimlessness of these or other kind of events, will certainly generate noisy parameters that in the long term may compromise the overall performance of the system.
For this reason, event-based semantic video summarization appears to be a feasible alternative.
In Song~et~al.~\cite{Song2016} the authors propose an event-centric video summarization method ({\em i.e.}, an approach not based on takes or key-frames).
The underlying method considers event detection based on trajectory analysis, and a random forest classifier to recognize abnormal deviations from the previously detected trajectories.
These detected abnormal events then enable a coverage algorithm that summarizes the complete relevant set of frames.

In most of these approaches, the main interest is to collect and represent relationships between different movements ({\em e.g.}, joints, poses) instead of collecting, analyzing, and visualizing large sets of sequential data.
This limits their applicability in large-scale projects like the ones required for I4.0. An example would be the combined analysis of several workers' activities over extended periods of time and in varying working conditions.
Some approaches take advantage of the advancements in virtual 3D worlds, CV, and data visualization.
So far none aims to produce an integrated solution that is able to provide all aspects required for ergonomic assessment of several workers, along large periods of time, and to summarize the analysis in sensible ways.
In this work, we propose an encompassing methodology that is able to:
\begin{itemize}
\item collect CV-based information related to workers' performance in real working sites, 
\item  evaluate this information regarding the most widespread ergonomic assessment methods, 
\item split the ergonomic analysis into domain tasks, determined by specific topic questions, in order to generate visualizations that fulfill the requirements, and 
\item present the results in a flexible and actionable dashboard that facilitates the most useful data manipulation operations to easily extract the relevant conclusions. 
\end{itemize}

\section{Tasks Abstraction and Requirement Analysis}
\label{sec:Requirement_Analysis}
% \textcolor{blue}{third sketch based on \cite{Munzner2009}}
% \subsection{Linking Goals to Questions}

We first perform a design study comprising a characterization of the problem domain, which in our case is an ergonomic hierarchical task analysis~\cite{Annett2003}.
As Meyer~et~al.~\cite{Munzner2009} suggested for real-world problem-driven studies, we initiated participatory design sessions with four domain experts.
% \textcolor{red}{how many, from where}.
In the sessions with the domain experts, we identified several analysis tasks for the exploratory analysis of ergonomics movement data.
The subdivision into tasks is similar to other commonly applied time-and-motion studies.
Splitting into tasks and the underlying question scheme are adequate to pinpoint and prioritize the circumstances that may be riskier. 
Our aim is to design a suitable interface to support ergonomic decision-making and thus improve workers' well-being.
The objects of analysis are ergonomic and angular distribution data and their temporal, similarity, quantity, and dependency relations as quantified through an automated video analysis.
The analysis task provides a dataset composed of tables describing the angular joint distribution per worker and per video frame.
These angles are calculated based on 3D CV body-joint inferences \cite{MassirisFernandez2020a}.
The underlying CV-based algorithms to estimate the ergonomic data have many parameters, for instance detection thresholds for filtering outliers, or acquisition-confidence factors if the video takes are not of good quality.
Extra filtering procedures may be needed due to other circumstances, like worker's occlusions or self-occlusions.
The REBA method uses the resulting joint angle information as the basis to calculate ergonomic risk.
It determines risk attributes with ordinal categories, where diverging low values represent a low ergonomic risk and viceversa.

The main focus of our approach is to support {\em a-posteriori} analysis tasks of time dependent scores obtained by one of the scoring schemes (REBA or RULA).
In an ergonomic evaluation case, the analysis aim might vary depending on what kind of information is most relevant to the stakeholders, and on the peculiarities of each evaluation task.
The ergonomic analysis begins with an overview of the time-and-motion study, which is afterwards decomposed into tasks at  desired levels of detail, in a similar vein to Schneiderman’s visualization mantra~\cite{Shneiderman2003}.
Then, tasks are sorted hierarchically, depending on various factors in light of the analysis purposes.
In particular, and after the interviews, we have distinguished six different evaluation tasks, which are decomposed into 17 basic questions that ergonomists have to address at different analysis stages (see Table~\ref{Tab:Ergonomic Task Analysis Steps}).
These questions were compiled from two sources: interviews with ergonomists about their data and analysis methods, and surveys of problems addressed in the literature \cite{Annett2003}.
The questions are the basis for defining the design goals of our visual analysis tool, and for conveying the relevant information and data relationships in a clear and distinguishable way.

\subsection{Task abstraction}

Defining the type of ergonomic analysis (T1) and the goals and evaluation criteria (T2), it is essential to highlight that these are context-dependent aspects.
An ergonomic analysis is generally part of the continuous improvement practices in enterprises and companies.
However, an on-premise ergonomic analysis may also be required due to regulation changes, direct expert or authority recommendations, the detection and determination of undesired situations, or as part of other corrective actions after contingencies or accidents. 
In this context, questions Q1 and Q2 clarify the focus of the ergonomic analysis, and questions Q3 and Q4 state the expected performance criteria.
In this initial phase, it is essential that our tool supports the user to understand what happened during the worker's activity, what should have happened in case something was undesirable, what might have happened in hypothetical contexts, and to estimate the rate and cost of failures.
Task T3 is concerned with adequate factorization of the analysis, decomposing the overall activity into smaller analysis units. 
Question Q5 focuses on the work-to-task division. 
Questions Q6, Q7, and Q8 point to the task triage. \change{The word {\em triage} is borrowed from medical parlance, in which a sorting by urgency determines the allocation of patients according to system priorities.
In the task triage, we sort each task in terms of input, output, protocols, records, value enrichment, and risk criteria.} 
During task T4, the requirement is to confirm the task decomposition, proposed goals, and performance benchmarks.
For this purpose, questions Q9 and Q10 intend to filter outliers or useless data by selecting an ideal task candidate, which represents the movements performed by a worker during a routine task.

Task T5 is concerned with understanding the actual ergonomic problems, in particular risky or potentially harmful movements or prolonged positions.
For this, we have to express if risky movements are associated with a specific task (Q11 and Q12), a body joint (Q13), or a body side (Q14). 
Finally, in T6 it is essential to arrive at conclusions, especially if a specific ergonomic improvement is required with some urgency.
In REBA, for instance, if a task is found to be of very high risk, an ergonomic intervention is demanded immediately. Our tool also provides specific support for this kind of actions (Q15).
Perhaps the most important question during the ergonomic analysis is Q16, which is how to mitigate ergonomic risk.
An expert should refer to the prevailing state-of-the-art or best practices to implement and validate (Q17) the \change{potential alternatives}.
The underlying purpose of Q16 and Q17 is to establish an impact measure.
This may have a translational effect in reducing accidents, diminish the incidence of WMSDs, retrain personnel, and overall save undesired ergonomy related-costs. 
Whenever possible, such an impact measure confirms the validity of the diagnosis and the proposed solutions.

\subsection{Requirements}
\label{subsec:requirements}

Once the analysis tasks were defined, we then elicited the following requirements from the ergonomy experts that participated in this study: 
%\vspace{-1.5ex}
\begin{itemize}%\setlength\itemsep{-0.1ex}
\itemsep0em 
    \item \textbf{R1:} Enable a fast path for an initial observation (T1).
	\item \textbf{R2:} Provide a way for observing the workers' movements and postures during several work cycles (T2).
	\item \change{\textbf{R3:} Quickly specify the most compromised postures, critical angle ranges, and highest force-load tasks (T3).}
    \item \textbf{R4:} Locate postures that are held during the longest period of time (T3). 
    \item \change{\textbf{R5:} Provide a way to find (or discard) atypical actions or joint-risk estimations (T4). }
    \item \textbf{R6:} Provide a way for visual and descriptive identification of the task and the movements performed (T1, T5).
    \item \change{\textbf{R7:} Provide means to easily locate frames or video portions with risky movements for workers’ retraining (T1, T6).}
    \item \textbf{R8:} Incorporate the REBA score tables which provide an action level with an indication of urgency (T3, T6)
    \item \change{\textbf{R9:} Provide comparisons of time-dependent scores for single and multiple joints (T5).}
    \item \change{\textbf{R10:} Quickly compare the risk distribution for each joint between the two sides of the body (T5, T6).}

\end{itemize}

\section{The {\tt ErgoExplorer} Design}
Based on the identified analysis tasks and the given requirements, we designed the {\tt ErgoExplorer} tool.
It provides means to quickly identify the focus required in a given analysis task, and to detect and understand relevant information within large amounts of information.
It offers an initial overview of the data with an adequate level of detail. The tool then allows to delve into specific subsets of the data ({\em e.g.}, when risky movements arise) and gather all the details relevant to the specific situation.
To make the {\tt ErgoExplorer} dashboard easier to interpret, we place a human body image in the center of the coordinated multiple views, which are then arranged around it, as shown in Fig.~\ref{fig:teaser}.
All the views are linked. If the user brushes in one view, the corresponding data items are highlighted in all the other views.
In the following subsections we present and motivate the rationale behind the design and use of the different views according to the tasks and requirements. 
We start with the description of visual encodings in {\tt ErgoView}, {\em i.e.}, the central view which is composed of several smaller views. 
We continue with the description of the {\tt ErgoTimeline}, and conclude the section with the\change{ chosen interaction design.
In order to illustrate the new techniques, we visualize ergonomic data from two video collections. The first dataset contains $15861$ video frames and $30$ corresponding ergonomic data attributes per captured video frame for a worker painting a wall (approx. 4.5 hours of video material). The second dataset has only $300$ frames and describes a person who does gymnastics.}

\subsection{Visual encodings in {\tt ErgoView}}
\label{ErgoView}
\begin{figure}[t!]
    \centering
    \includegraphics[width=\columnwidth]{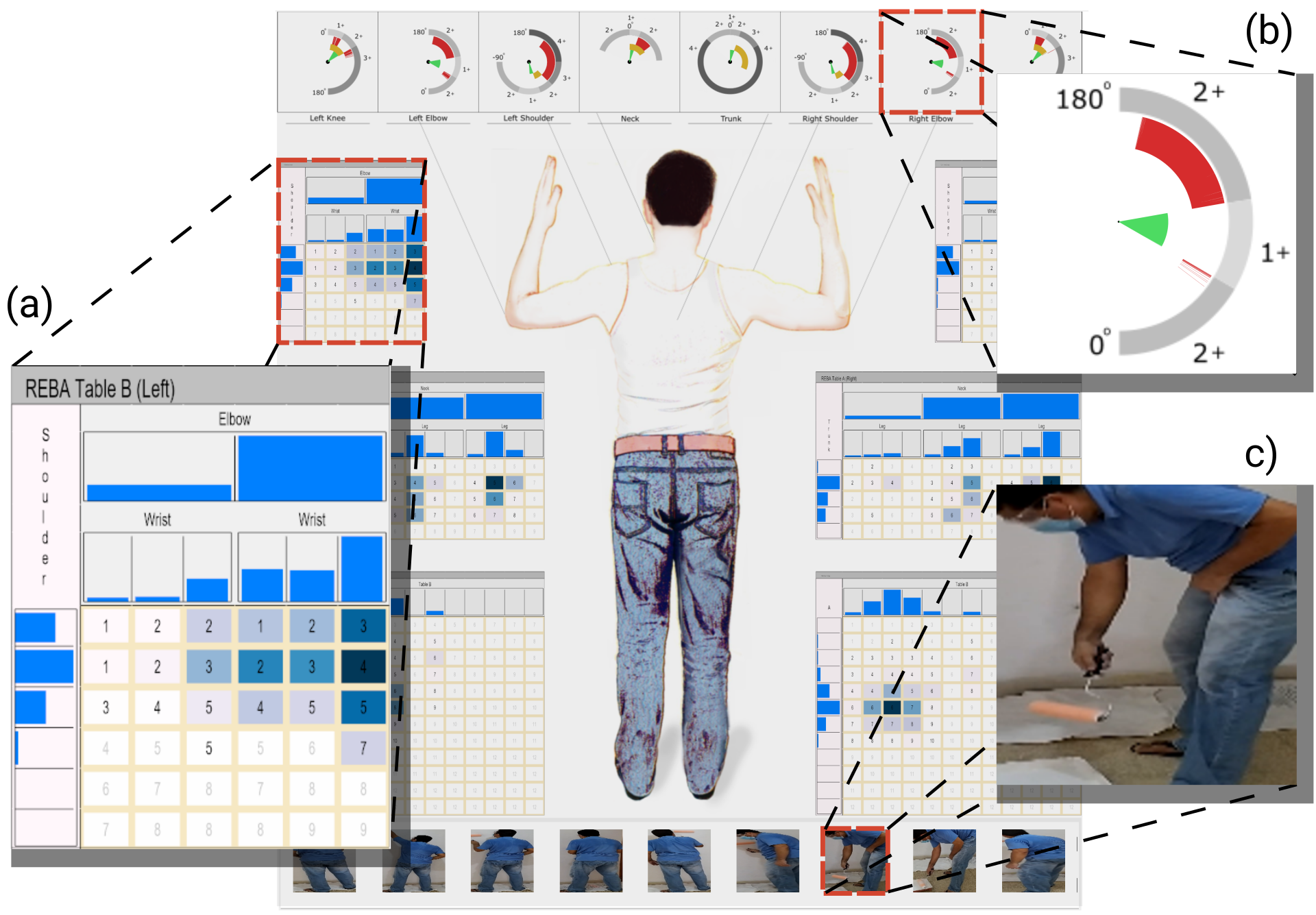}
    \caption{{\tt ErgoView}. %The spatial relation of the three sub-views to the ErgoView is emphasized.
    \textbf{(a)} The {\tt REBA Table B} is maximized, which shows the data attribute Score for the wrist, lower arm, and upper arm of the left body side (see Fig.~\ref{fig:REBATableDesign} for details on the design of the augmented score tables).
    \textbf{(b)} The {\tt ErgoGauge} is used to visualize the entire range of measured values for the corresponding joint angle; here, the right-shoulder view is maximized.
    \textbf{(c)} The {\tt ErgoMovements} view addresses the requirements (such as R5, and R7) to enable image(s) or video preview, for example, to inspect a time window around a specific pose \change{(\textbf{R1})}. 
    }
    \label{fig:ErgoViewExploded}
\end{figure}

The main component, {\tt ErgoView}, is the central view for the exploration, which includes other views arranged around a human body image (see Fig.~\ref{fig:ErgoViewExploded}).
This arrangement addresses different design requirements, presents known information appropriately and, according to the experts' opinion, engages, and facilitates the specific analysis tasks.
Besides the human silhouette which serves as an orientation landmark, the {\tt ErgoView} contains {\tt REBA Tables}, {\tt ErgoGauges}, and {\tt ErgoMovements}.
The well-known {\tt REBA Tables} represent a link to the conventional way in the domain of risk assessment, and they are required by the domain experts (\textbf{R8}).
{\tt REBA Tables} are arranged around the picture of a human body, where the body silhouette is seen from behind. This orientation makes it easier for ergonomists to quickly associate the scores in the tables positioned on the right side with the corresponding joints on the right side of the human body, and analogously for the left side.
The same is true for the {\tt ErgoGauges}, which are placed in a row above the silhouette. They support a detailed analysis of ergonomic risk for all body joints (\textbf{R9}). 
In addition to the proper left and right placement, a line also links each gauge to the corresponding body joint, making the visual connection even more explicit.
Finally, at the bottom we have the {\tt ErgoMovements} view, which makes it possible to include further sources of information into the analysis, such as images and video (\textbf{R7}).
Each of the three components of the {\tt ErgoView} is explained in detail in subsections below.
The {\tt ErgoView} layout proved to be very practical according to the experts, and it is used as a starting point in the analysis.  
%In the initial analysis phase it is used without additional views that we have in our ErgoExplorer, to clarify the focus of the ergonomic analysis and define the expected performance criteria. 

% 
% \textcolor{red}{Describe different panes of the view: circles, tables, silhouette, and movie line. Justify designs with tasks from the last section. If needed add additional tasks.}

\subsubsection{\tt {REBA Tables}}
\label{REBAs tables}

At the beginning of an analysis experts need to quickly reveal the spread of potentially risky movements in the data (\textbf{R1}).
Our view of {\tt REBA Tables} is designed to support this basic requirement.
% In general, the REBA method separates the body into joint sections (relative to planes of motion) and provides a hierarchical scoring system based on the aggregation of individual joint scores.
In particular, it uses three scoring tables (A, B, and C) that reflect the division of a body into sections, including the upper arm, lower arm, wrist, neck, trunk, and legs.
Tables A and B show raw data attributes, while Table C receives scores from tables A and B as input.
\change{Values from Table C with an added activity score constitute the final result of a posture evaluation, which is an action level with an indication of the severity of the assessment.}

Because {\tt REBA Tables} are still widespread in the application domain, we display them (\textbf{R8}), but in an interactive and cumulative way  (\textbf{R3}).
In our approach, posture scores are computed automatically using CV methods as explained before. A vast number of scores for an extended period of time can be calculated and deployed for analysis automatically.
%Although it is possible to adapt our system to work in an online manner (the data becomes available in real-time in a sequential manner, one REBA posture score at a time), in this paper, we opted for an a-posteriori analysis.
We display all scoring tables at once, three for the left and three for the right body side (\textbf{R10}).
Moreover, we propose to augment all tables to allow ergonomists to compare time-dependent scores for single and multiple joints (\textbf{R9}).
Our design relies on histograms and a heatmap to show time-dependent data while retaining the primary purpose of {\tt REBA Tables}.

\begin{figure}[t!]
    \centering
    \includegraphics[width=\columnwidth]{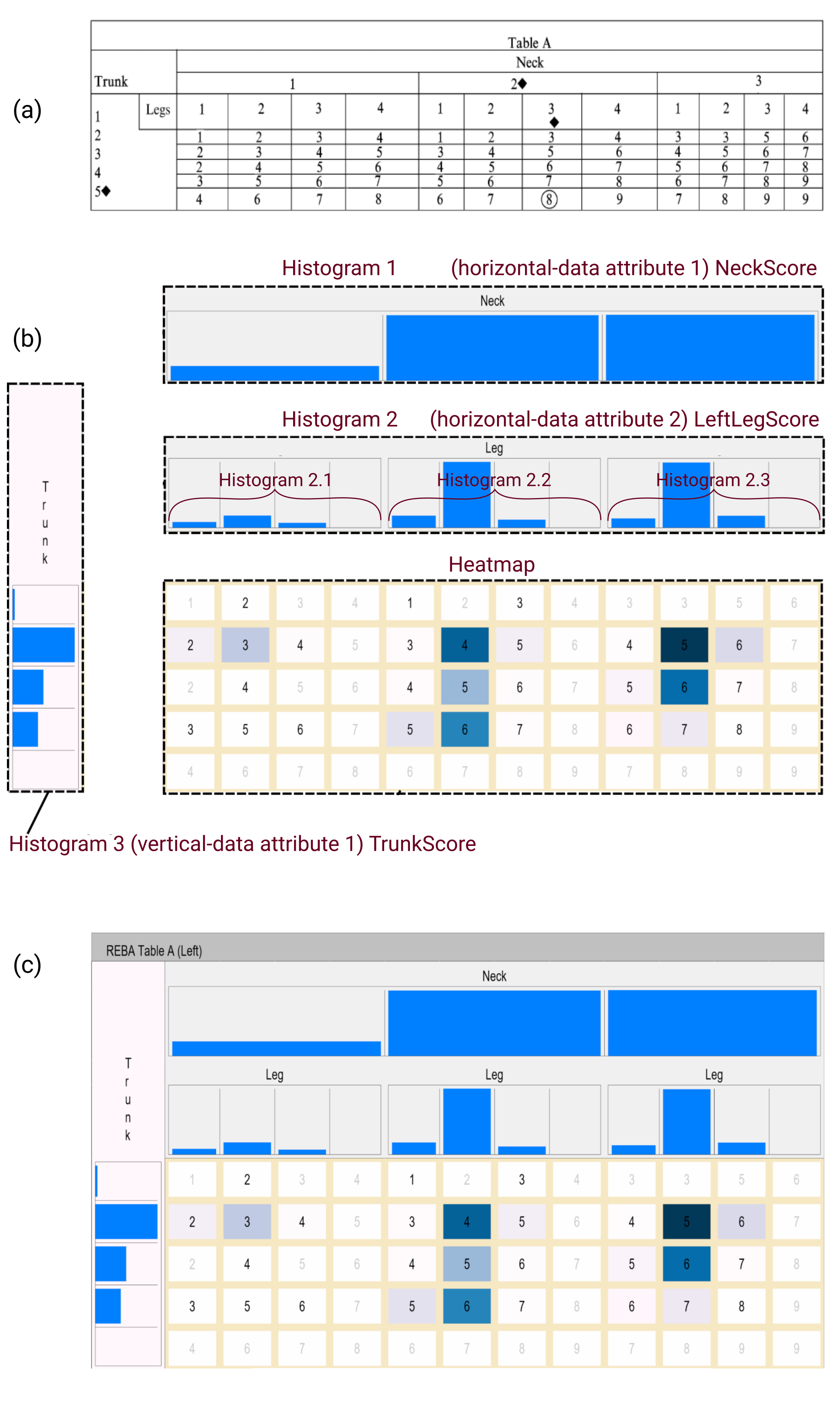}
    \caption{Design details of the {\tt REBA Table A}. \textbf{(a)} The original table design as published by McAtamney and Hignett ~\cite{McAtamney2000}. \textbf{(b)} The main parts of the augmented table. \textbf{(c)} The proposed Table A which is interactive and includes vertical and horizontal histograms, and a heatmap \change{(\textbf{R8})}. 
    }
    \label{fig:REBATableDesign}
\end{figure}

We briefly explain the original table design.
Since all tables share similar designs, we will use Table A as an example.
The three data attributes used in this table are TrunkScore, NeckScore, and LegScore.
The original design of Table A is shown in Fig.~\ref{fig:REBATableDesign}(a).
Note that {\tt TrunkScore} goes from 1 to 5, {\tt NeckScore} from 1 to 3, and {\tt LegScore} from 1 to 4.
The posture scores for each attribute are pre-entered in the corresponding cells. 
Ergonomists can quickly select one value for each joint based on their observation of the worker's posture. 
In this figure, a diamond icon indicates the assigned score (2 for the neck, 3 for the legs, and 5 for the trunk).
An established hierarchical scoring scheme is used, which means that in order to calculate the final posture score, ergonomists must go from the top data attribute (Neck). 
Depending on the score given there, they mark the score in the level below (Legs) and then find the intersection with the third attribute (Trunk) to read the derived score, which is 8 in this case.
The same procedure is repeated for Table B and for Table C.
Because each cell in a table can hold only a single scalar value, the original {\tt REBA Tables} are not appropriate for analyzing a collection of observations.
For this reason, we adopt the basic table design, but allow each cell to hold more than one value. 
We encode the resulting value in color, and the table becomes a heatmap.
We also add marginal histograms for rows and columns.
To explain our design, we use terminology that differentiates between horizontal and vertical data attributes---this relates to rows and columns in the tables, respectively, but also to how the histograms in the tables are oriented.
For example, Fig.~\ref{fig:REBATableDesign}(b) shows four horizontally oriented histograms (one for horizontal-data attribute {\tt NeckScore}, and three for horizontal-data attribute {\tt LeftLegScore}) and one vertically oriented histogram (for vertical-data attribute {\tt TrunkScore}).
The numbers of bins in the histograms in the first level, {\em i.e.}, for the horizontal\_data\_attribute\_1, and vertical\_data\_attribute\_1, correspond to the number of posture scores that can be assigned to the respective joint. 
In the second level, {\em i.e.}, for the horizontal\_data\_attribute\_2, the number of histograms corresponds to the number of bins in the histogram that is one level above. 
Also at this level, the number of bins in each histogram corresponds to the number of scores for the respective joint.
Since at the first level there is only one histogram the total number of data items is the sum over all bins. 
However, at the second level, the total number of data items is the sum of all bins of all histograms.
Here, each histogram has as many items as contained in the bin of the first-level histogram located at the top of a second-level histogram.
{\tt REBA Tables} do not use more than two levels.
However, our approach allows for more horizontal or vertical levels in a table ({\em e.g.}, the RULA tables are commonly designed with two vertical levels).
The height of a bin in a histogram indicates the frequency of the corresponding joint score in the data, while a heatmap uses color intensity to indicate the frequency distribution of the computed posture scores.
We opted for a heatmap because we wanted to keep the numerical values of the overall REBA score for each table cell as well. 
If we use a single posture measure, instead of showing many measures, then on each level there is only one populated bin in the histogram, and in the heatmap only one cell is used. This would be the same as looking at the original table depicting only single scalar values.
The dataset shown in the example case has more than $15 000$ rows, {\em i.e.}, unique posture measurements. 
As histograms and heatmaps represent aggregated visualizations, depicting such an amount of data is not a problem.

%Histograms combined with a heatmap proved to be a good way to quickly reveal the spread of potentially risky movements in the data.
%In order to take into account the lateralization in each activity, we made two instances for each of the tables A, B and C,  corresponding to the left and right side of the body.

\subsubsection{\tt ErgoGauge}
\label{Ergogauge}
\begin{figure}[t!]
    \centering
    \includegraphics[width=\columnwidth]{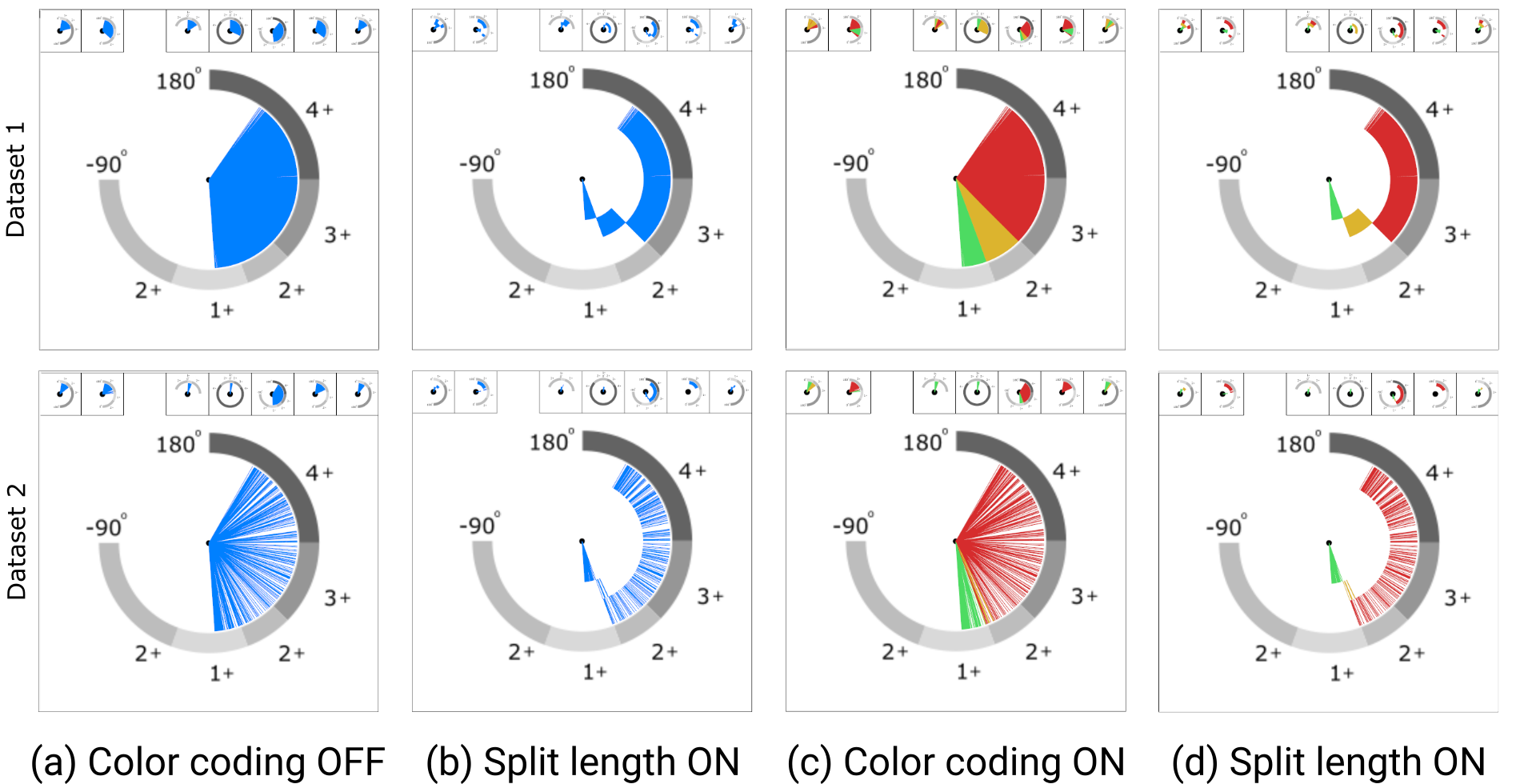}
    \caption{Four possibilities to depict the actual range of the recorded angles using the {\tt ErgoGauge} are shown. \textbf{(a)} and \textbf{(b)} use only one color (information) while \textbf{(c)} and \textbf{(d)} use three different colors (the usual heat map palette). \textbf{(d)} uses color and different line lengths. Since angles are only indirectly related to posture scores, this can be very useful for discovering deviating or inappropriate working postures that can affect the workers’ performance or their health.}
    \label{fig:ErgoGaugeDesign}
\end{figure}

%\begin{figure}[ht!]
%    \centering
%    \includegraphics[width=\columnwidth]{figures/ErgoGauge1.png}
%    \caption{ErgoGauge supports ergonomics analysis by displaying the whole range of possible angles (according to the scoring scheme used) and the actual range of the recorded angles.}
%    \label{fig:ErgoGauge}
%\end{figure}
The REBA method defines a safe posture for each joint in terms of the most desirable joint angles, and our {\tt REBA Tables} support a whole-body ergonomic risk evaluation.  
The requirement \textbf{R3} describes the need to support a detailed ergonomic-risk distribution-analysis for each joint. 
A safe posture, {\em e.g.}, for the elbow, means the forearm position is within an angle between $60$ and $100$ degrees concerning the vertical axis of movement.
As we have angles to visualize, we decided to use a radial layout.
We show a single measurement angle as a radial line.
Since data are angle measurements of human body joints, we display only a part of the circle corresponding to the valid range of a particular joint movement.
In this way, a quick connection between the actual and valid joint movements is established.
Fig.~\ref{fig:ErgoGaugeDesign} shows the basic design of the view.

Linking an angle to its assigned score is also an important aspect that we want to convey in this visualization.
Since angles are only indirectly related to posture scores, we use color coding to communicate the movement risk for each joint angle.
The {\tt ErgoGauge} classifies and locates each joint's estimated risk according to the ergonomic angle using the traffic light palette (red, yellow, and green), with red indicating the worst score.
We choose red, yellow, and green after discussions with domain experts although these colors are not distinguishable for persons with impaired color vision. 
In such a case, the color palette can be modified accordingly.
Varying the length of segments (without coloring them) is another option to reduce cognitive demand and help ergonomists to concentrate on analyzing and comparing different angle scores and related angle ranges (see Fig.~\ref{fig:ErgoGaugeDesign}(b)). 
While this choice was appreciated by ergonomists, it was found that, in rare cases, different posture scores may be assigned very similar or the same joint-angle value. Other parameters such as load may affect the given score. 
In this particular case, the combination of the color and length channel to encode the joint score proved to be better in emphasizing such an unusual posture (\textbf{R5}). 
A corresponding {\tt ErgoGauge} example is shown in Fig.~\ref{fig:ErgoGaugeDesign}(d).
Without color coding, the {\tt ErgoGauge} shows only the possible range of joint movements.
All three design choices are shown in Fig.~\ref{fig:ErgoGaugeDesign}.
\change{The top row shows $Dataset1$ with fifteen thousand entries compared to the bottom row with $Dataset2$ and a few hundred entries.}
Using only the {\tt ErgoGauge} by itself, it is hard to know how many items are within a certain range.

\subsubsection{\tt ErgoMovements}
\label{ErgoMovements}
\begin{figure}[t!]
    \centering
    \includegraphics[width=\columnwidth]{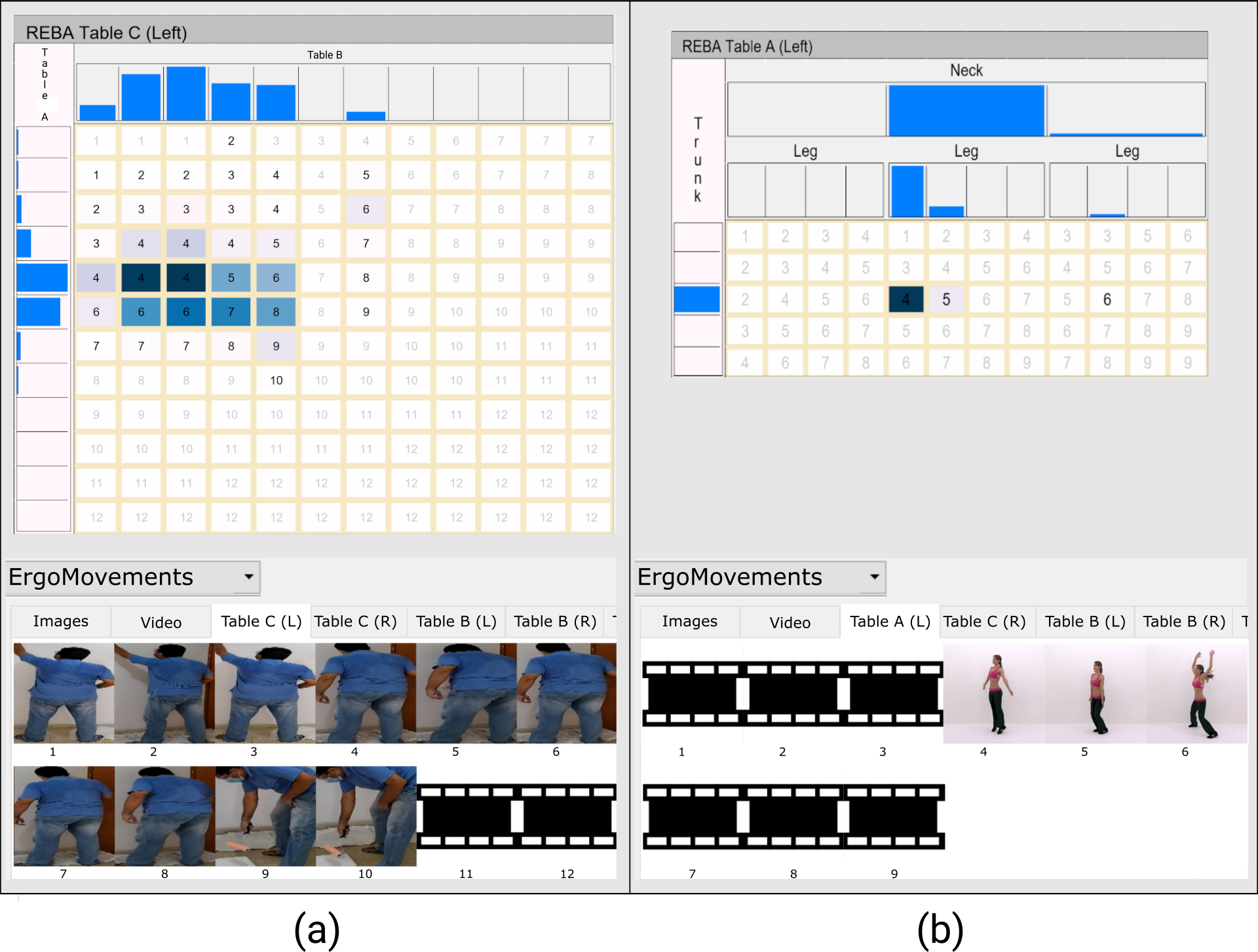}
    \caption{\textbf{(a)} For each of the scores computed in Table C, an image is automatically selected that gives visual feedback to the expert about the worker's action performed. In the shown case, the worst scores are nine and ten, and the related action is dipping a roller into a container with paint placed on the floor \change{(\textbf{R7})}. \change{\textbf{(b)} Table A for $Dataset2$ is sparse, however, risk assessments for legs, neck, and trunk are most of the time within an acceptable range.}}
    \label{fig:ErgoMovements1}
\end{figure}

There is a general trend towards an automatic monitoring in workplaces.
Ergonomy experts will be provided with detailed ergonomic data to conduct a thorough a-posteriori analysis. 
However, visual data will still be the most important information in their work. 
One of the requirements has been that {\tt ErgoExplorer} should include means to analyze videos and posture images during the assessment task. 
This can be accomplished with the {\tt ErgoMovements} view, which provides standard video playback options (\textbf{R7}) and also supports further requirements, including \textbf{R1}, \textbf{R2}, \textbf{R3}, \textbf{R5}, \textbf{R7}, and \textbf{R6}, as explained in the following.
\change{Compared to the pencil-and-paper-based REBA method, {\tt ErgoMovements} helps users to create} a better mental representation of the analyzed data by linking numerical values in the views together with actual workers' movements.
For example, by examining the quantitative and qualitative data presented in the {\tt REBA Tables} and displayed in the {\tt ErgoView}, the ergonomist can quickly conclude the seriousness of the situation and whether to react immediately.
In order to help analysts perceive the depicted information more efficiently (especially regarding \textbf{R5}), we provide options to display different sets of pre-selected images (see Fig~\ref{fig:ErgoMovements1}).
Images are one of the attributes in our datasets, and each image is associated with a specific time point.
The user can select any of the related data tables shown in the {\tt ErgoView}, to display the corresponding set of images.
Each of the images relates to its corresponding REBA score. There is an option to quickly switch between the tables to gain insight into the worker's actions in relation to the tables' scores. 
% An example is displayed in Fig.~\ref{fig:ErgoMovements1} B, where REBA Table C gives an overview of twelve different scoring levels, but in the data we analyzed the two worst ratings ($11$, and $12$) were not registered, so two images were not displayed.

We have considered different ways of selecting representative images since quite different postures can result in the same overall score in the table. 
In working conditions where actions are repeated cyclically (as is our case), the experts mentioned that any image of a group with the same score is a good representative of the whole group.
In this case, they identified the relevant task to establish a relationship between the results presented in the {\tt REBA Tables}, the complexity of the work, and the related risk factors to which the worker is exposed.
{\tt ErgoMovements} helps to clarify the observed workers' ergonomics data in the context of their original work environment (\textbf{R2}, \textbf{R6}, \textbf{R7}).
Moreover, {\tt ErgoMovements} can show examples of unsafe actions as well as good practices previously executed during the workday (\textbf{R3}, \textbf{R5}, \textbf{R7}), which in turn supports the reduction of ergonomic risks.
For instance, workers who are at a high ergonomic risk undergo retraining sessions.
During this retraining, {\tt ErgoMovements} depicts representations of the currently performed movements and postures (\textbf{R7}), highlighting the aspects that need to improve, and also the progress to achieve safer working practices. 
%Visual search is also a way to provide documentation and valuable knowledge about rare events, especially in non-routine tasks (Q10).
% Brushing in the linked views is reflected in the ErgoMovements, allowing visual and descriptive identification of the task and the movements performed (T5). 

% \textcolor{red}{El frame viewer presenta la imagen resumida del trabajador evaluado, para ello se hacen hacen recortes de la minima caja circundante del trabajador en el video orginal. La seleccion en diferentes graficas se ve reflejada en el Frame Viewer, lo cual permite una identificacion grafica y muy descriptiva de la tarea y movimientos realizados. Ademas, el frame viewer ayuda a descartar movimientos o estimaciones atipicas.}
% Visual observation is a way to provide documentation and valuable knowledge about rare events, especially in non-routine tasks.

\subsection{\tt ErgoTimeline}
\label{Ergo Timeline}
\begin{figure}[t!]
    \centering
    \includegraphics[width=\columnwidth]{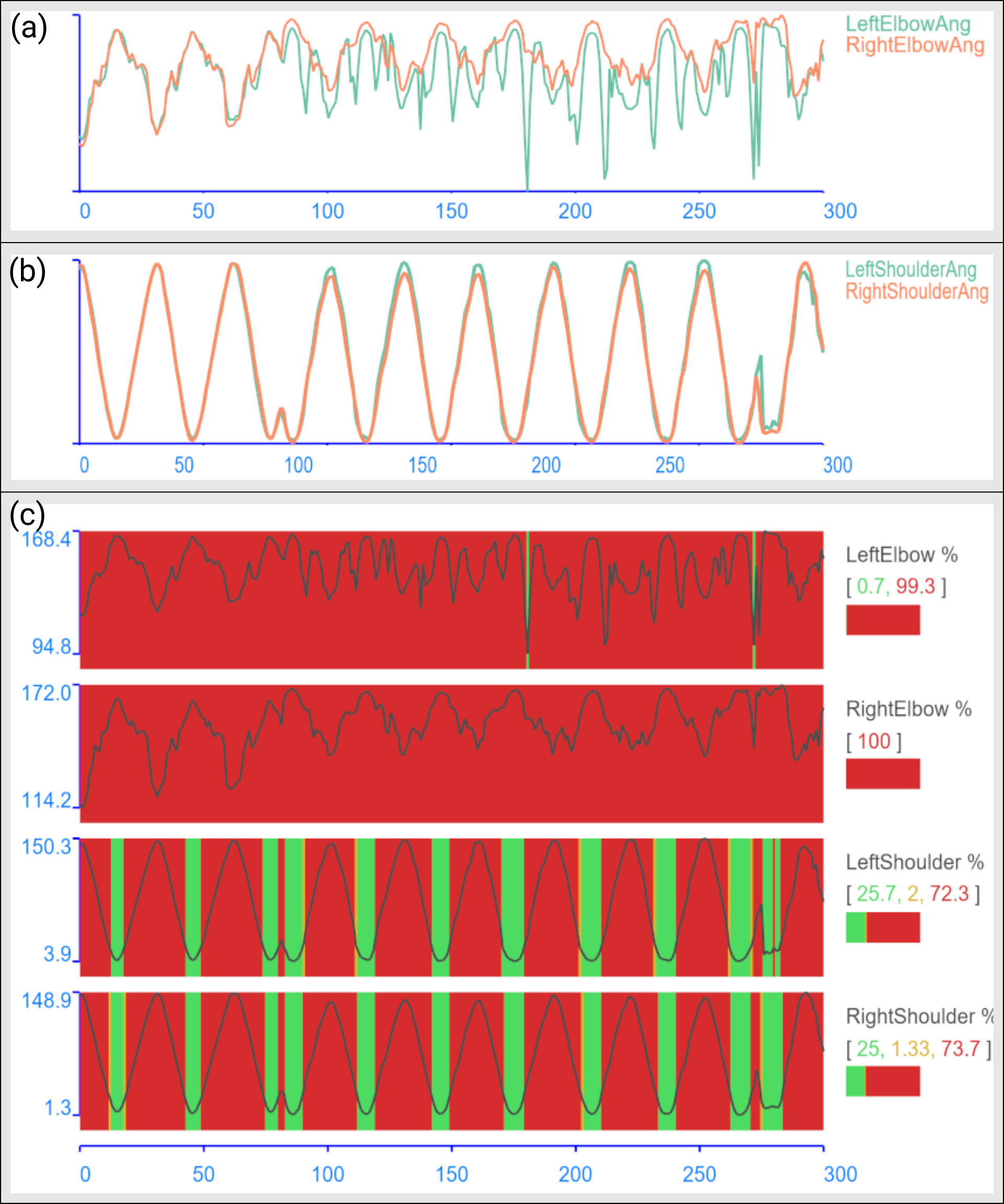}
    \caption{\change{{\tt ErgoTimeline} to depict angular values over time for gymnastics movements ($Dataset2$). \textbf{(a)} and \textbf{(b)} compare angles for the left and right elbow and for the left and right shoulder, respectively. \textbf{(c)} simultaneously compares angles \& risks of the elbows and shoulders. Different body joints operate in different angle ranges Each vertical axis is scaled according to the range of the corresponding data attribute (\textbf{R9}).}}
    \label{fig:ErgoTimeline}
\end{figure}

%The ErgoTimeline supports visualization at different levels of detail, depending on the analysis tasks.
%Concerning the task decomposition chart (T3), 
\change{The {\tt ErgoTimeline} shows the distribution of joint angles and their risks over time (see Fig.~\ref{fig:ErgoTimeline}).
{\tt ErgoTimeline} analysis (T3) depicts an action's repetitions, duration, and other time-related aspects. This is practical for analyzing routine {\em a-posteriori} instead of on-premise working contexts.}
% The ergoTimeline shows over time the distribution of joint angles and their risks (see Fig.~\ref{fig:ErgoTimeline}).
% {\tt ErgoTimeline} analysis (T3) deals with repetitions, duration, and other time related aspects of an action. 
% This leads to the analysis of a routine, which is different to on-premise working contexts. 
Usually, the selection of routine levels and a work-to-task partitioning (work sampling) are made by firsthand inspection, which is time consuming and lacks inter-rater reliability~\cite{Schwartz2019, Gold2006}. 
Based on the tasks that make up the work at hand, five levels of decreasing routinization can be distinguished~\cite{Gold2006}: i) a constant task with a predictable work cycle; ii) various cyclical tasks; iii) a mixture of cyclical and non-cyclical tasks; iv) a single non-cyclical task, and v) multiple non-cyclical tasks.
The {\tt ErgoTimeline} assists experts in assessing as to whether the analyzed work has some level of routinization and how an action can be split into smaller parts. 
Moreover, it directly supports tasks T3 and T4, the decomposition of actions, and its validity. 
It also supports the evaluation of repeatability and facilitates the elimination of outliers. 
The main concept of this view is to depict the joint angles by a line chart.
In this way, the analysts see the values as a function of time, and can easily spot cycles, repetitions, and irregularities (\textbf{R2}, \textbf{R5}).
We show data for several joints in a single chart to support a visual correlation analysis (\textbf{R9}).
Color coding and labels are used to distinguish between joints in this case.
Finally, there are limits which indicate a risky or unhealthy value for each joint (\textbf{R7}). 
We depict the current limit values as vertical lines colored in green for non-risky and in red for risky postures. 
{\tt ErgoTimeline} also superimposes the curves on colored background (\textbf{R3}). An example is given in Fig.~\ref{fig:ErgoTimeline}(c) to examine joint angles in detail.
As limits are given per joint only, in case of comparing several joints a common vertical axis becomes impossible. In this case the vertical axis is split across parts of the view. 
An advantage of {\tt ErgoTimeline} is that we do not rely on regular work cycles. 
We provide a scoring scheme for joint angle-movements created to handle static, dynamic, or unstable postures based on REBA. 
% If the ergonomics expert faces multiple non-cyclical tasks, the time spent per task cannot be estimated. 

%\begin{figure}[t!]
%    \centering
%    \includegraphics[width=0.8\columnwidth]{figures/tmp1.png}
%    \caption{Comparison of angles for the left and right elbow.}
%    \label{fig:Angles_timeline_300_Elbow}
%\end{figure}

%\begin{figure}[t!]
%    \centering
%    \includegraphics[width=0.8\columnwidth]{figures/tmp2.png}
%    \caption{Different body joints (here shoulders and elbows) operate in different angle ranges.}
%    \label{fig:Angles_timeline_300_Elbow_shoulder} 
%\end{figure}

%\begin{figure}[t!]
%    \centering
%    \includegraphics[width=\columnwidth]{figures/ErgoTimeline2.png}
%    \caption{\change{Long-term observation of the work task ($Dataset1$).}}
%    \label{fig:scores_timeline_big_data_shoulder}
%\end{figure}
The measured joint angles are not the only attributes for calculating the posture scores, but they are still crucial for a more detailed ergonomy analysis, {\em e.g.}, to determine how an injury occurred.
\change{In Fig.~\ref{fig:ErgoTimeline}(a) we observe the effect of fatigue and lateral unbalance. At the beginning both shoulders move evenly and as the worker's right shoulder gets tired, the left shoulder begins to execute more abrupt movements. In Fig.~\ref{fig:ErgoTimeline}(b), we see at the end of the sequence a posture correction made by the gymnast that stands upright. 
Finally, it is possible to inspect how a variation in the joint angle affects the estimated risk score with the overlap display, as depicted in Fig.~\ref{fig:ErgoTimeline}(c). The {\tt ErgoTimeline} indicates a high risk exposure for elbows and cyclic movements in the shoulders}.
% \textcolor{red}{Describe differentiate modes, and what are they good for. Again use tasks from above and add tasks if needed.} 
% introduce ergonomic score time line. Which tasks are covered by this view? visualization at different levels of detail, depending on analysis tasks.

% Mod 1 Angles-Only Timeline: Advantages (used in Task X)/ Disadvantages

% Mod 2 Scores-Only Timeline: Advantages (used in Task Y)/Disadvantages

% Mod 3 Angles-and-Scores Timeline: Advantages (used in Task Z)/Disadvantages

% \textcolor{red}{I don't understand well this: 
% The advice of an action (" do x ") instead of goal-directed communication (" achieve state x ") while interpreting feedback could contribute to their underestimation \cite{Annett2003}.}
% As a result, our development will aid in describing an ergonomic state.

%\subsection{Additional views}
In addition to the specifically designed views described above, we also allow for other standard views if needed. 
During the analysis tasks, parallel coordinates and scatter plots are often used to explore correlations between values.
The layout is fully configurable, and Fig.~\ref{fig:teaser} shows the setup preferred by the domain experts, though other settings are possible as well.
% We provide a suggested layout, but the views are freely configurable, and experts can change the layout according to specific needs.

% ErgoExplorer aims to support ergonomists in analyzing data to answer questions and provide measures that can reduce the time an operator spends in a non-ergonomic posture.
% In this respect, conventional views, such as scatter plot matrix or parallel coordinates, are also deployed.
% TODO Explain the usage of additional views
% We provide a suggested layout, but the views are freely configurable, and experts can change the layout according to specific needs.

\subsection{Interaction Design}
\label{Interaction}
\begin{figure}[h!]
    \centering
    \includegraphics[width=1.0\columnwidth]{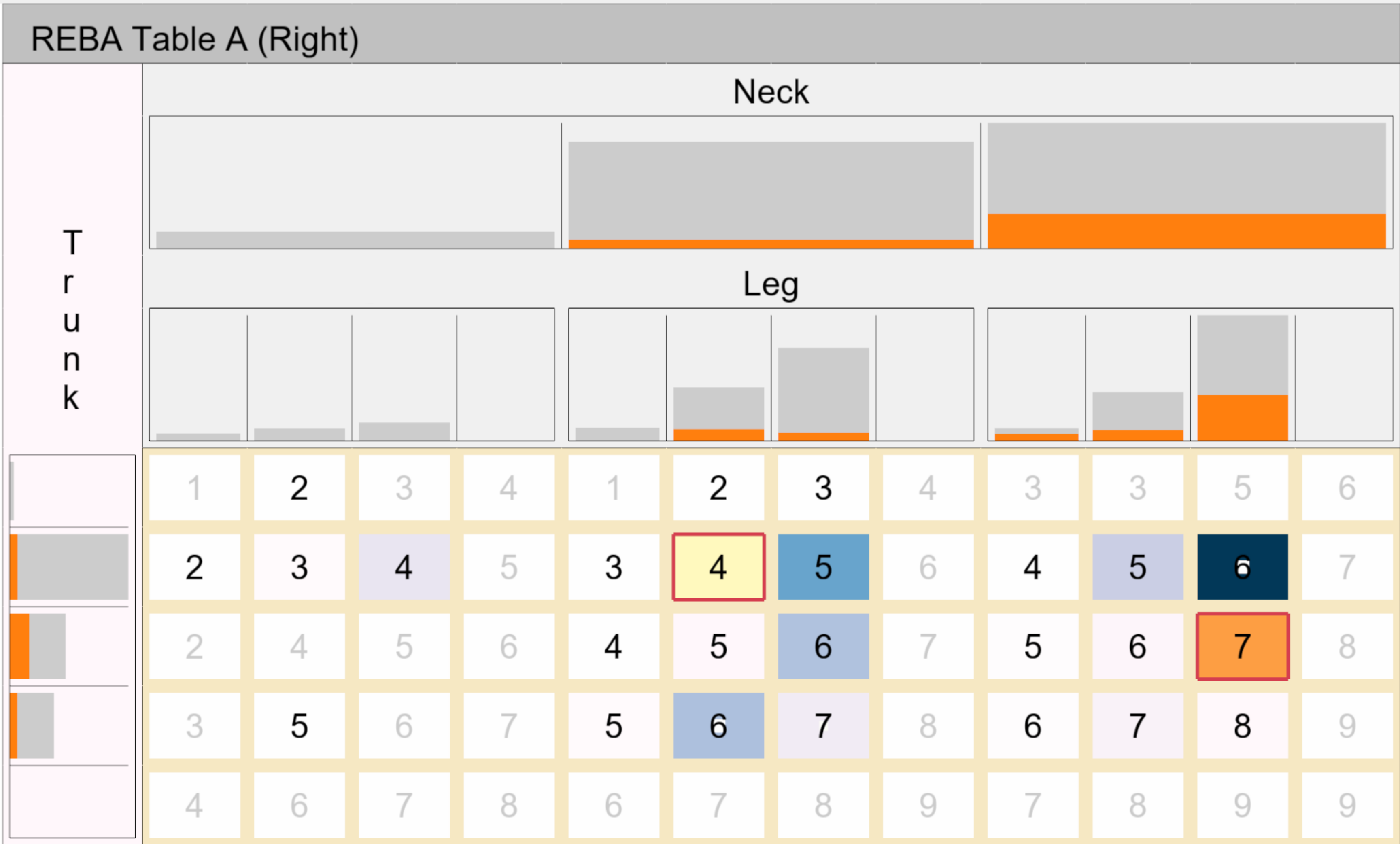}
    \caption{Brushing in {\tt REBA Tables}. Two cells (highlighted by red rectangles) are selected in a heatmap (\change{R8)}.}
    \label{fig:Table_A_HeatmapBrushed}
\end{figure}

\begin{figure}[h!]
    \centering
    \includegraphics[width=0.8\columnwidth]{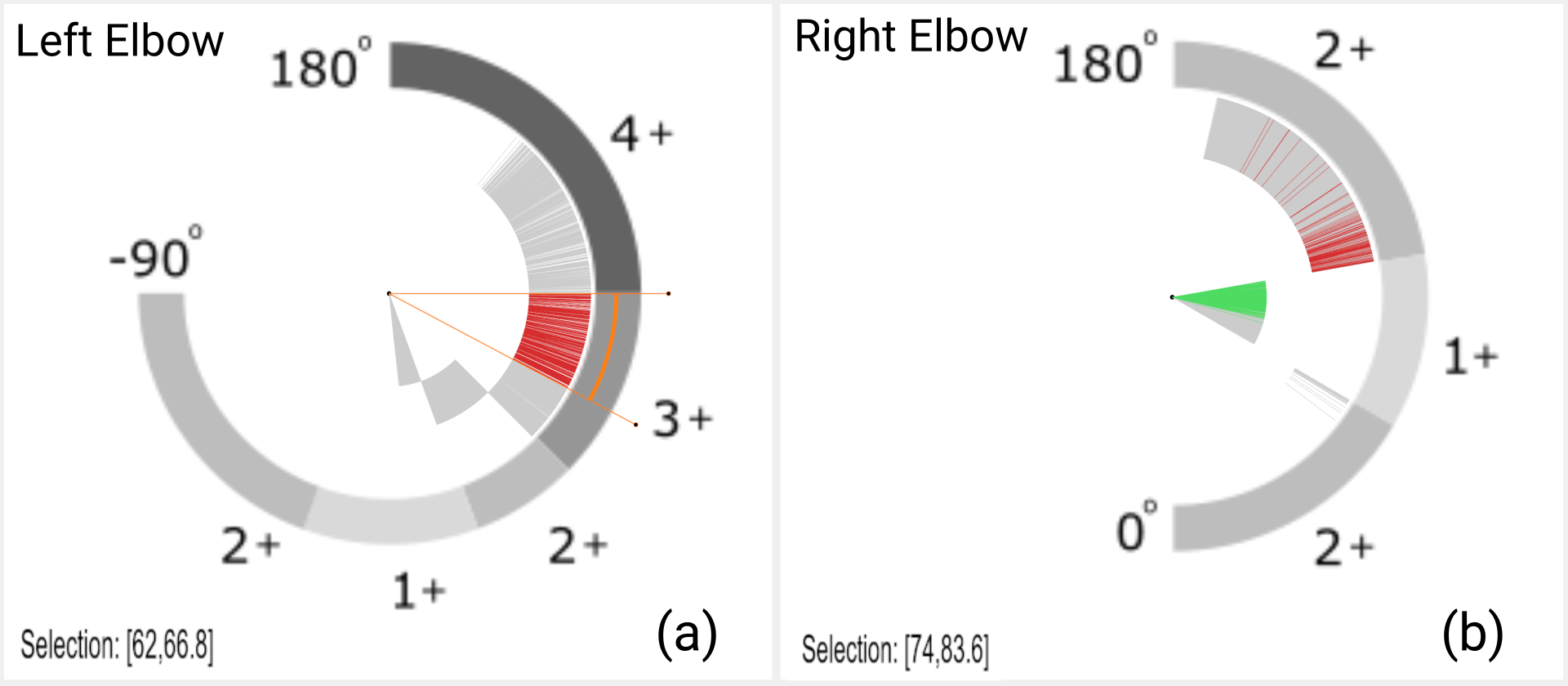}
    \caption{Brushing in {\tt ErgoGauge}. \textbf{(a)} A brush was created to select the right-shoulder angles in the range from $62$ to $66.8$. \textbf{(b)} Corresponding radial lines are highlighted for the right elbow in the linked {\tt ErgoGauge} \change{(\textbf{R10})}.}
    \label{fig:ErgoGaugeBrushed} 
\end{figure}

Actionable views are essential for some requirements ({\em e.g.}, {\bf R3}, {\bf R6}). % to gain insight into causes, relationships, and dependencies.
For instance, brushing facilitates selecting the critical task to be ergonomically evaluated (Q9).
In all views, we support very simple types of data selection, {\em i.e.}, through mouse clicks. Where needed, we added more advanced brushing operations, including composite brushing and an angle-brush that selects a user-defined angle on a circle in the {\tt ErgoGauge} view.
Concerning the {\tt REBA Tables}, the user can select a single bin in a histogram or make a composite brush by selecting several histograms with the mouse.
If a user selects one or more cells in the heatmap, we regard this action as adding new data items to the same brush and not as creating a new brush and adding it into a composite brush (see Fig.~\ref{fig:Table_A_HeatmapBrushed}). 
Ergonomists appreciated this realization because, in most cases, they select a couple of cells (posture score values). Then they observe the relations in other views \change{to generate hypotheses or valuable clues about the possible causes of actual ({\bf R6}) or potential ergonomic risks ({\bf R8}) and suitable feedback recommendations ({\bf R7}}). 
With {\tt ErgoGauges} we support the analysis of joint angles that deviate from a safe posture.
We implemented a brush that follows the radial layout of its parent view, as shown in Fig.~\ref{fig:ErgoGaugeBrushed}.
We support a quick selection of all data items that share the same joint score by enabling the user to click on the outer ring of the ErgoGauge. 
% {\em i.e.}, over one of the gray sectors (each sector corresponds to the score noted on the side).
Also, the user can adapt the brush handles or enter the values in numeric fields for exact positioning. %(the values are textually depicted on the lower left corner).
More than one brush can be specified, allowing the user to combine different angle ranges. This may be needed to create a reference stage for a risk-distribution analysis.

Analyzing the time-dependent data is of high priority for ergonomists.
To support this requirement, we implemented a range brush in the {\tt ErgoTimeline}. 
The brush can be placed at an arbitrary position on the horizontal axis, and the user can change the brush's extent and position at any time.

\begin{figure}[ht!]
    \centering
    \includegraphics[width=\columnwidth]{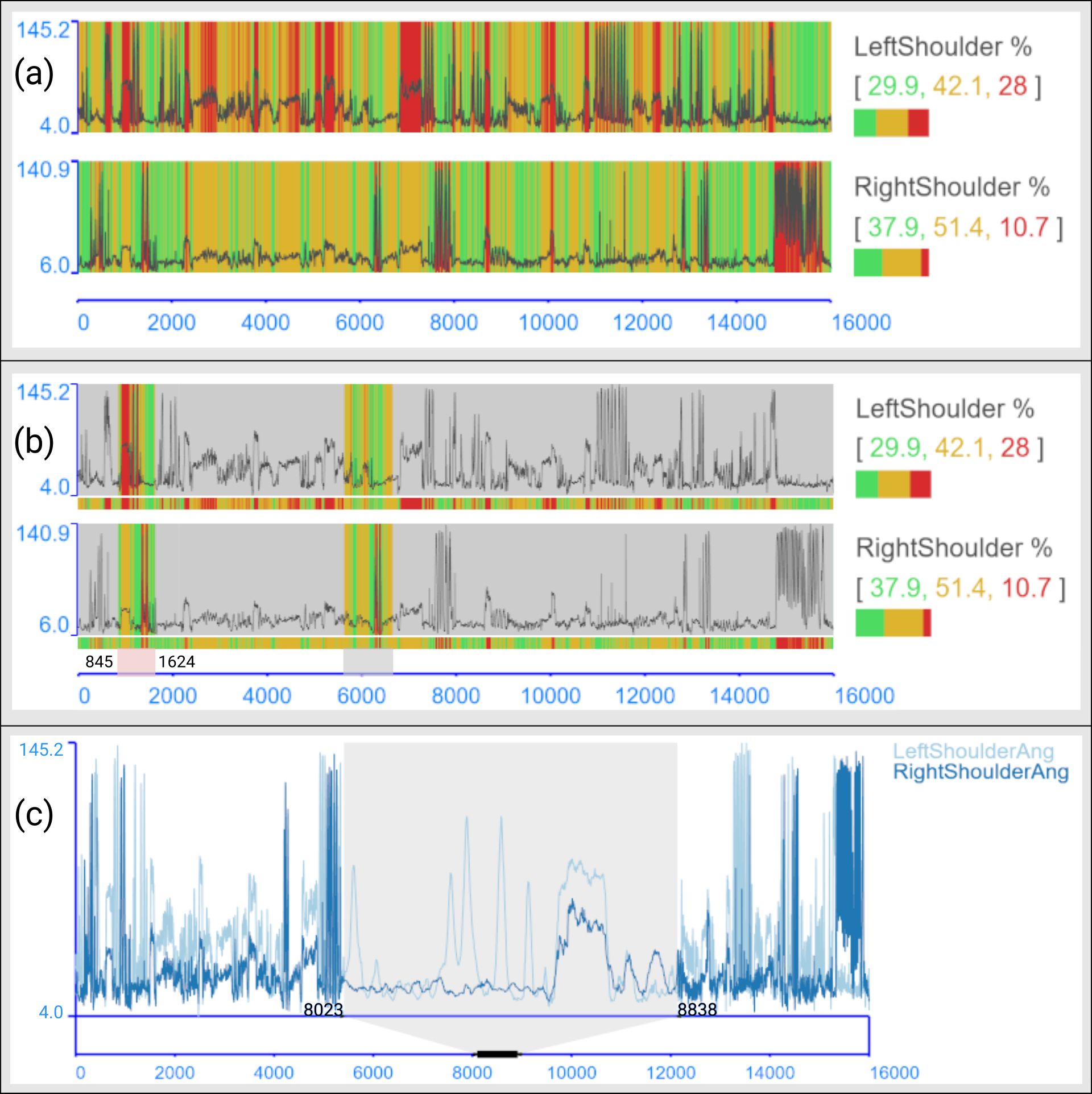}
    \caption{Interaction in {\tt ErgoTimeline}. \textbf{(a)} The
    angular values over time for worker's movements ($Dataset1$) \textbf{(b)} Two brushes were created to select distinct ranges on the angles time-line for long-term observation of the wall painting task. The range values for the active brush (pink) are also shown. \textbf{(c)} Part of the timeline was zoomed in to analyze the disparity in movement between the right and left shoulder \change{(\textbf{R9})}.} 
    \label{fig:brushing_ErgoTimeline}
\end{figure}

The path of curves or the color-coded results in the {\tt ErgoTimeline} are clearly visible for visualizing data with a small number of time steps. 
\change{However, in some scenarios, posture measurements must be taken over a prolonged time, and as a consequence, fine details in the {\tt ErgoTimeline} may be lost. As shown in Fig.~\ref{fig:brushing_ErgoTimeline}(a) it is impossible to perceive all the curves' fine-detail changes.
Multiple brushes can be created to support time-interval comparisons, as shown in Fig.~\ref{fig:brushing_ErgoTimeline}(b).
We have implemented a magnifier (zoom slider) as a details-on-demand option that can be adjusted in two ways.
First, the users specify the range on the temporal, {\em i.e.}, horizontal, axis that the magnifier should enlarge. Then, they decide how much of the view space is dedicated to display the magnified data (see Fig.~\ref{fig:brushing_ErgoTimeline}(c)). In this way it is possible to create fine-detail brushes on the temporal axis.
Second, it is possible to set up the magnifier to work in the other direction, {\em i.e.}, to compress a large portion of the data onto a tiny part of the display.}
Our method allows the user zooming in and out to filter outliers in the work sequences (Q10) and to compare values obtained in the other views.

\section{Discussion}

As {\tt ErgoExplorer} was modeled in a multidisciplinary effort, the complete design cycle received significant feedback from expert ergonomists. 
In the design, a fast access to situations and focus on information where REBA risk scores are high, has been a vital goal. 
%In particular, 
\change{Based on the analysis tasks and questions (see Table~\ref{table_ttt}), {\tt ErgoExplorer} has been able to simplify the experts' activities and to quickly pinpoint all interventions if they are immediately required ({\bf R1}).
The {\tt REBA Tables} are especially useful in analyzing tasks with real or potential hazards. They are able to summarize the accumulated risk distribution along complete work cycles with different time granularities ({\em i.e.}, simple task repetitions, complete tasks, complete working days, etc.).
}

\change{Also, as mentioned in Sect.~\ref{subsec:requirements}, determining and segmenting work cycles are lengthy and (sometimes cumbersome) procedures.
In this regard, the {\tt ErgoTimeline} presents an easy understandable representation of the ergonomic risks along time. It allows the users to discern the work cycles and all of the movements that characterize them ({\bf R2}).
This is especially relevant to pinpoint a single working cycle that can be taken as a representative of all the other cycles. The goals are to localize inadequate working postures ({\bf R3}) or movements that repeat or extend for an inappropriately long duration ({\bf R4}), and to pinpoint outliers or anomalies ({\bf R5}).
The traditional determination of work cycles is prone to subjective visual-interpretation biases. Therefore, it has been appreciated by the users to easily link the numerical ergonomic risk estimations in the {\tt ErgoTimeline} ({\em e.g.}, peaks) to viewing frames ({\bf R6}) in the actual video capture where the associated events occurred.}

\change{Comparing the same movement performed across different tasks, has been another analysis task facilitated by the combination of {\tt ErgoTimeline} with viewing frames.
This possibility was quite well received by the users. It allows the experts to retrain workers concerning movements where they are exposing themselves to undue risks ({\bf R7}), or otherwise to show them exemplars on how to correctly perform a task.
It is common knowledge in the ergonomy literature that good practices are adopted faster if workers are able to see for themselves in the very moment when they are making a wrong movement sequence~\cite{NIPUNDEBNATH2021} ({\bf R7}).}

\change{As a noteworthy characteristic of {\tt REBA Tables}, the final users needed almost no explanation since they are familiar with understanding the meaning of a plain traditional {\tt REBA Table}.
For instance, if most values in the {\tt REBA Table C} are concentrated in values above 8 \cite{McAtamney2000}, this means that the ergonomic risk is high and the current working activity requires an intervention by a specialist ({\bf R8}).}

\change{Experts} considered valuable to present angular estimations directly in the {\tt ErgoGauge} view with a 180\textdegree \hspace{1 em} rotation \change{This generates a mirror-like opposing visual between both sides of the body}, instead of showing them in a separate user-interface widget. The users can remain focused on the main purpose of the visualization ({\bf R10}). 
In addition to the presentation of the most frequent conditions, ergonomists are also interested in detecting irregular or undesirable body movements, and in identifying where and  when these movements arise.
This type of analysis is not easily feasible without a visualization tool like the one presented here. A manual inspection of the data at the required level of detail would be a daunting task.
Our interactive visual analysis tool supports this and other complex tasks by utilizing the linking\&brushing mechanism of the coordinated multiple views (CMV).
For instance, the user brushes in one of the views a region where angular deviations can be considered risky or inadequate. The selected movements are highlighted in all the other views, facilitating the remaining analysis tasks ({\bf R10}).
A zoom slider was added to simplify searching fine details and filtering with immediate response. 
After initial demos and presentations of our approach to the domain experts, several modifications were incorporated following their comments and suggestions ({\bf R9}).

\begin{figure}[ht!]
    \centering
    \includegraphics[width=\columnwidth]{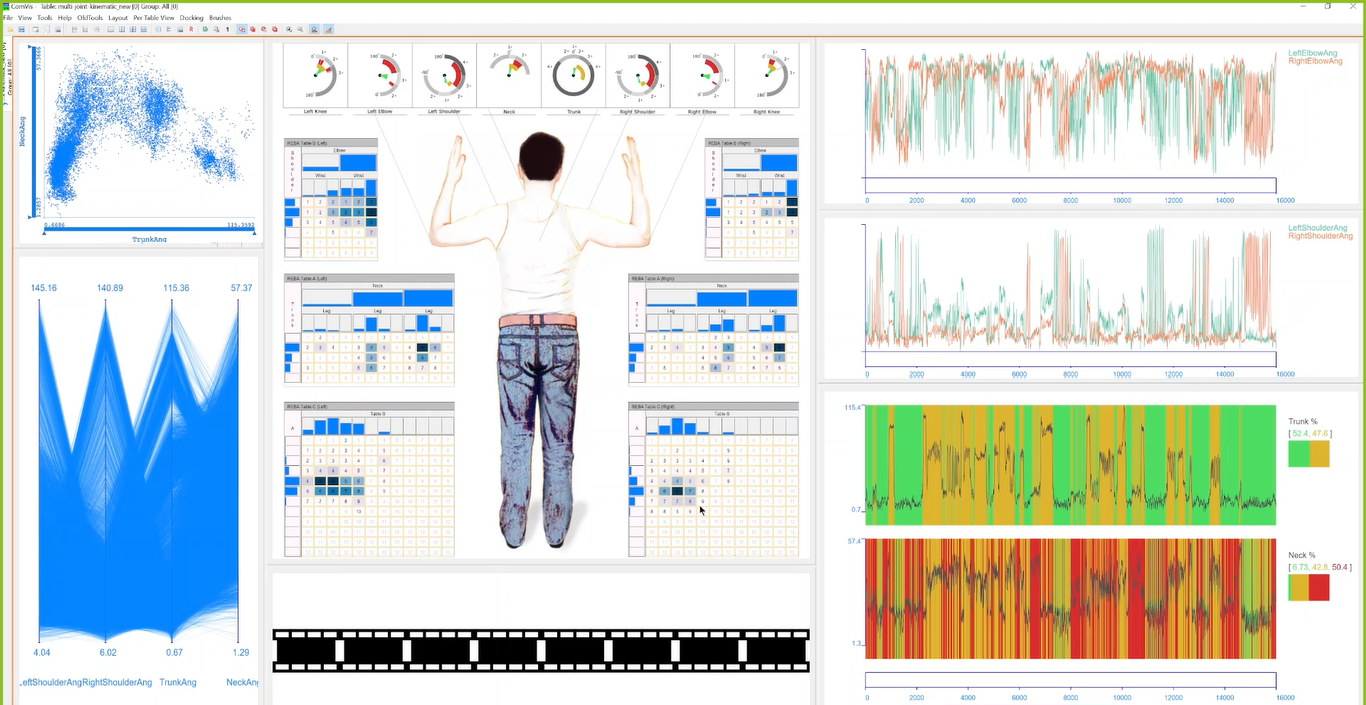}
    \caption{Main view of {\tt ErgoExplorer} \change{(\textbf{R1})}}
    \label{fig:video1a}
\end{figure}

\begin{figure}[ht!]
    \centering
    \includegraphics[width=\columnwidth]{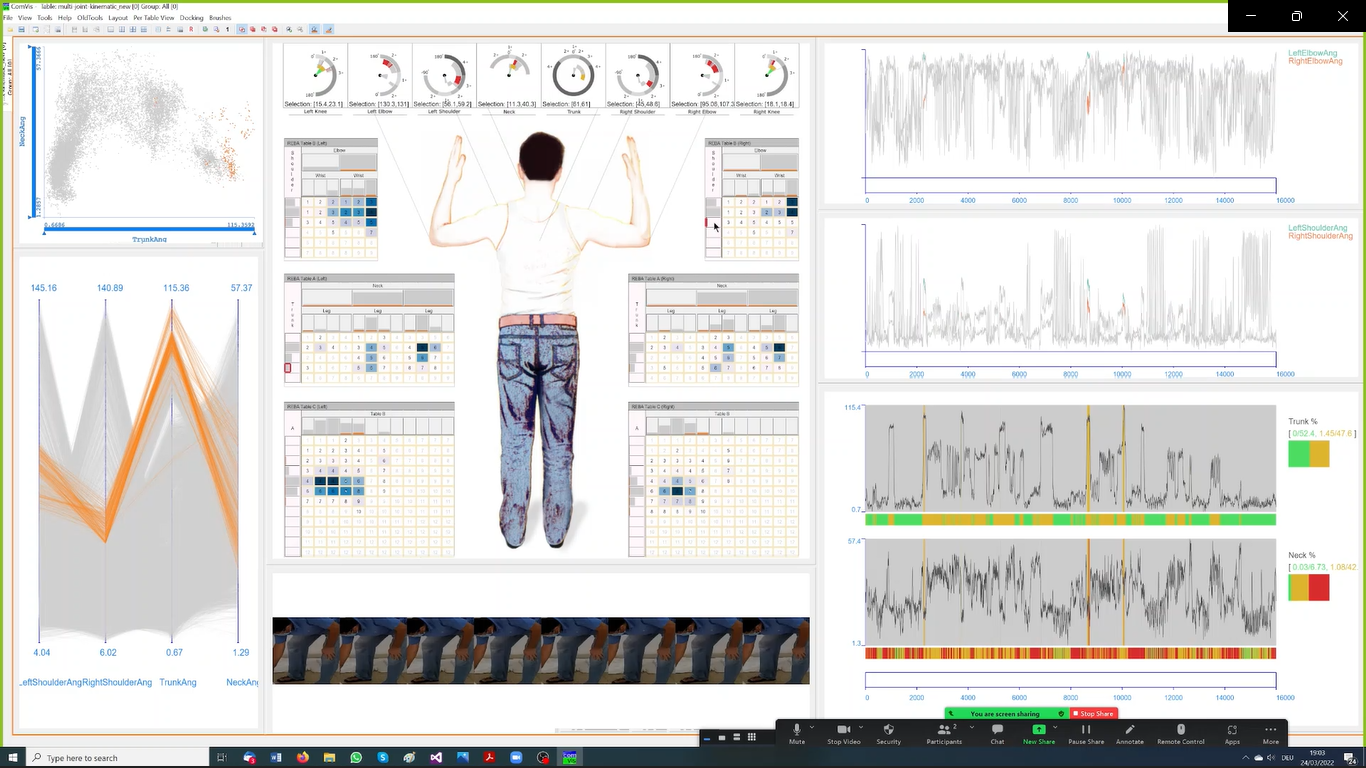}
    \caption{\change{Riskiest body positions are highlighted through brushing and depicted in the frame view (\textbf{R3})}}
    \label{fig:video1b2}
\end{figure}

\begin{figure}[ht!]
    \centering
    \includegraphics[width=\columnwidth]{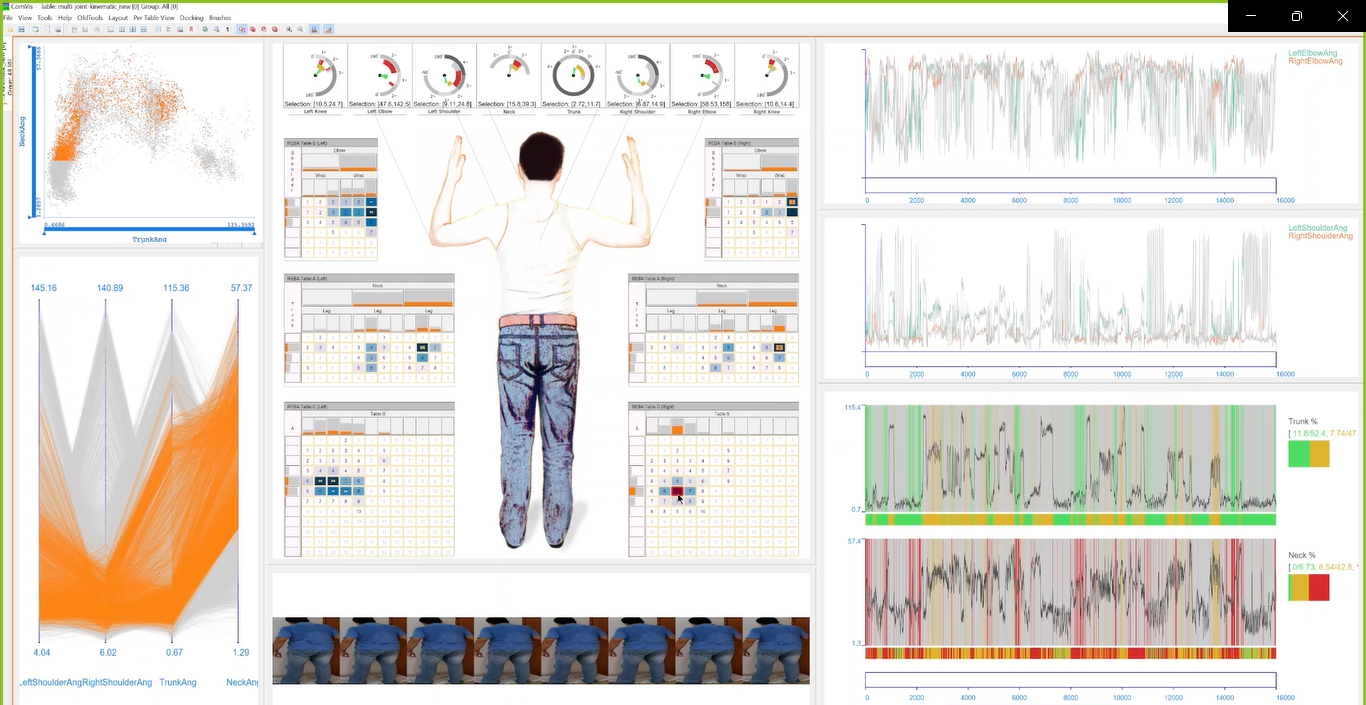}
    \caption{Repetitive postures are highligthed through brushing (\textbf{R4})}
    \label{fig:video1c}
\end{figure}

\begin{figure}[ht!]
    \centering
    \includegraphics[width=\columnwidth]{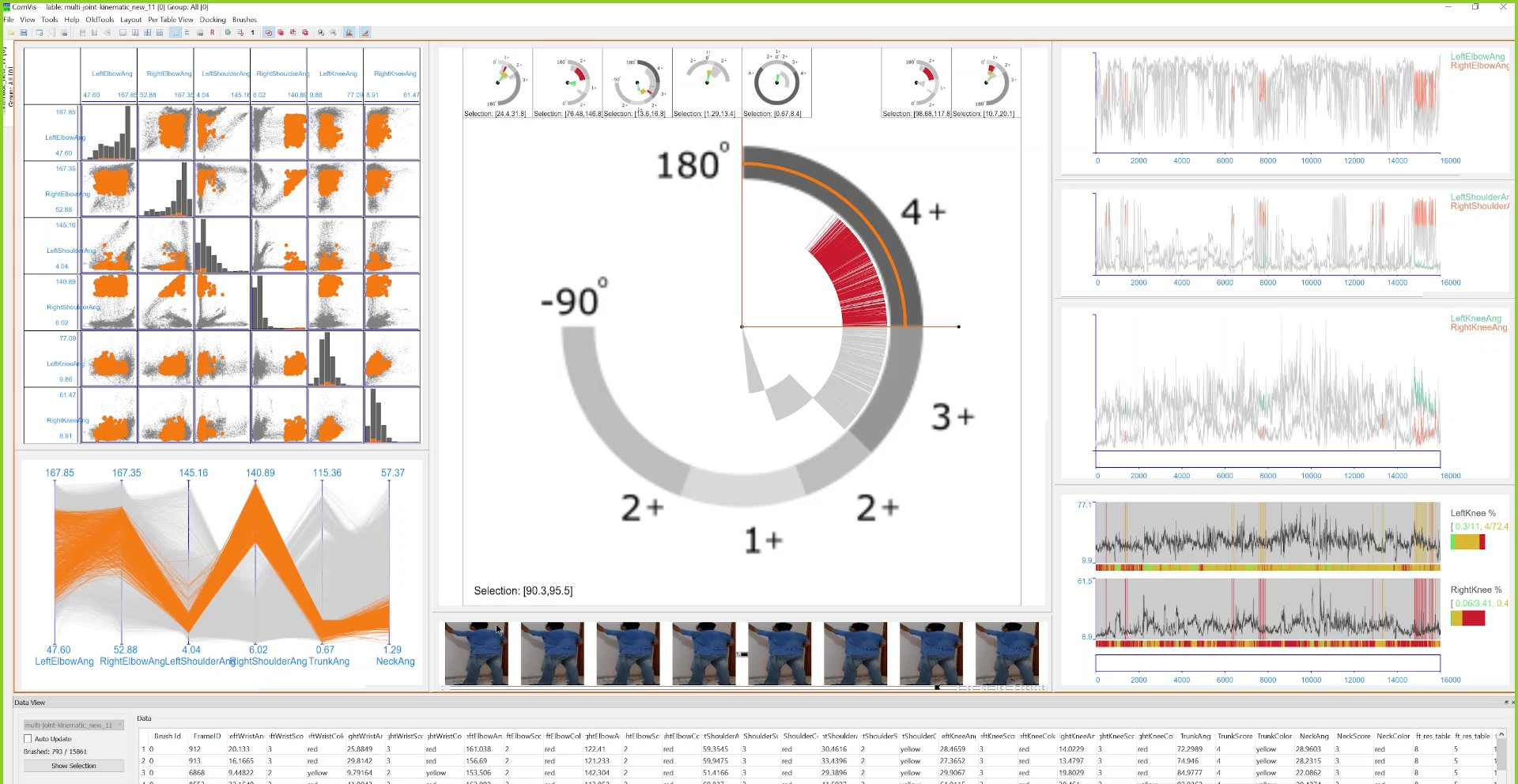}
    \caption{\change{Selection of right-shoulder angles with high risk (\textbf{R10})}}
    \label{fig:video2}
\end{figure}

The accompanying video shows a representative example of a frequent use case.
% Next, we illustrated a typical use-case of our visual approach.
Initially, we represent responses to the ergonomic questions Q1, Q2, and later to questions Q8, Q9, and Q10. 
With the main view of {\tt ErgoExplorer} (Fig.~\ref{fig:video1a}), a user can define the focus of the ergonomic analysis. Typically it will be a routine inspection, but may also be a more pressing scenario where improvements are needed immediately.
Following question Q8, the ergonomists can prepare a task triage, where they prioritize the inspection of high risk (Q11), more repetitive (Q6), or abnormal movements (Q9 and Q10).
{\tt ErgoMovements} provides access to the brushing results. Pictures containing the riskiest body positions are shown in Fig.~\ref{fig:video1b2} and the maximal REBA risk score is given in Fig.~\ref{fig:teaser}.
Additionally, the linkage to other views allows the users to focus on particular body segments that generate the riskiest body positions.
The histograms in the tables highlight repetitive postures (Fig.~\ref{fig:video1c}). 
Fig.~\ref{fig:video2} shows the results of brushing high-risk angles of the right shoulder through outlier detection on the {\tt ErgoTimelines}.

% replaced by the teaser to cite it!
% \begin{figure}[ht!]
%
%     \centering
%     \includegraphics[width=\columnwidth]{figureexamples/risky_posture.png}
%     \caption{Maximum REBA risk score}
%     \label{fig:video1b}
% \end{figure}

% \begin{figure}[ht!]
%     \centering
%     \includegraphics[width=\columnwidth]{figureexamples/scatterplot_Moment.png}
%     \caption{relation between neck and trunk}
%     \label{fig:video3}
% \end{figure}

% \begin{figure}[ht!]
%     \centering
%     \includegraphics[width=\columnwidth]{figureexamples/scatterplot_Moment(2).png}
%     \caption{relation between neck and trunk}
%     \label{fig:video3}
% \end{figure}

\section{Conclusion}
% conclusion

We present {\tt ErgoExplorer}, an approach for the interactive analysis, visualization, and interpretation of ergonomic risks in video sequences. It is based on the REBA scores of joint positions and angles as derived with computer vision techniques. 
Our aim is to develop useful visualization techniques for managing time series data of ergonomic information in conjunction with the video sequences. The purpose is to facilitate assessment tasks to pinpoint risky situations, repetitive tasks, and other undesirable conditions that may arise in workplaces.
We propose a taxonomy of ergonomic evaluations based on a task analysis by generating the key questions that -- according to the knowledge elicitation of the domain experts' activities -- best describe the ergonomic analysis process.
The complete ergonomic assessment cycle comprises the finding of adequate answers to the key questions following the proposed methodology.
Based on the collaboration with the domain experts, several future research directions are being considered.
The most salient is to incorporate the RULA ergonomic risk evaluation (and perhaps other, newer assessments) into our analysis.
This would lead to a more comprehensive and complete pipeline for ergonomic analysis.
Also, a freely available {\tt ErgoExplorer} installation is expected to facilitate large-scale cooperative research activities in ergonomics, through the collection and assembly of much larger datasets with which more elaborate statistical analyses will be feasible.
We initiated interchanges with the Colombian Academic Network of Ergonomics, and also with several ergonomists in Argentina, to provide them with free access to {\tt ErgoExplorer}. This will allow us to access richer (properly anonymized) datasets, and enhance the usefulness of the proposed visualization and interaction approaches. 

\section*{Acknowledgments}

% We are grateful for the support of ergonomic experts: Prof. Beatriz Tsukamoto, Prof. Sandra Liliana Joaqui, Dr. Martha Torres, and Prof. Ayda Cáceres from the Colombian academic ergonomics network.

The authors thank all the workers and experts who helped in the development of this research, with a special mention to the Occupational Health and Safety Service at the Universidad Nacional del Sur ({\tt https://uns.edu.ar}), and  Prof. Beatriz Tsukamoto, Prof. Sandra Liliana Joaqui, Dr. Martha Torres, and Prof. Ayda Cáceres from the Colombian Academic Network of Ergonomics ({\tt  https://scergonomia.com.co/race}). VRVis is funded by BMK, BMDW, Styria, SFG, Tyrol and Vienna Business Agency in the scope of COMET - Competence Centers for Excellent Technologies (879730) which is managed by FFG. This research was partially founded by the National Council for Scientific and Technical Research of Argentina (CONICET); and the SGCyT-UNS (grant 24/K083).

\typeout{}
\bibliographystyle{abbrv-doi}
\bibliography{library}

\end{document}